\documentclass[runningheads]{llncs}

\usepackage{eccv}

\usepackage{eccvabbrv}

\usepackage{graphicx}
\usepackage{booktabs}

\usepackage[accsupp]{axessibility}  

\usepackage{hyperref}

\usepackage{orcidlink}

\graphicspath{{figures/}}

\usepackage{pifont}
\usepackage{booktabs}
\usepackage{multirow}
\usepackage{rotating}
\usepackage{algorithm}
\usepackage{algpseudocode}
\usepackage{capt-of}
\usepackage{url}
\usepackage{mathtools}
\usepackage{subcaption}
\usepackage{colortbl}
\usepackage{nicematrix}
\usepackage{enumitem}
\usepackage{placeins}

\newcommand{\Fref}[1]{Fig.~\ref{#1}}

\begin{document}

\title{InterCMDM: Block-Causal Diffusion for Autoregressive Human Interaction Generation}

\titlerunning{InterCMDM}

\author{Qing Yu\orcidlink{0000-0001-6965-9581} \and
Kent Fujiwara\orcidlink{0000-0002-2205-6115}}

\authorrunning{Q.~Yu and K.~Fujiwara}

\institute{LY Corporation, Tokyo, Japan\\
\email{\{yu.qing,~kent.fujiwara\}@lycorp.co.jp}}

\maketitle

\begin{abstract}
Text-conditioned human interaction generation must capture both long-range temporal causality within each individual and tightly coupled coordination between partners. Existing interaction diffusion models typically denoise full sequences using bidirectional attention, which obscures causality and hinders streaming and long-horizon generation. Autoregressive alternatives enforce causality but often suffer from temporal drift, leading to coordination degradation and unstable interaction dynamics over time. We propose InterCMDM, a block-causal latent diffusion framework for autoregressive two-person interaction generation. InterCMDM introduces a Dual-Stream Causal Diffusion Transformer that maintains separate causal streams for each person while modeling inter-person dependencies via unified dual-stream attention with multi-task attention masks. These masks unify interaction modeling within a single attention mechanism and support diverse coordination behaviors, including simultaneous actions, reactive responses, leader--follower dynamics, and independent motion. By training a single model across these mask configurations as a form of data augmentation, InterCMDM enables controllable interaction generation by simply selecting the desired attention mask at inference time. Finally, a block-wise diffusion objective enables stable latent rollout over long sequences without repeated decode--encode cycles. InterCMDM achieves state-of-the-art performance on InterHuman and Inter-X, improving text--motion alignment, realism, and long-horizon continuity.

\keywords{Human Interaction Generation \and Motion Synthesis \and Autoregressive Diffusion Models}
\end{abstract}

\section{Introduction}
Text-conditioned human interaction generation aims to synthesize coordinated two-person motion that matches a natural-language description. Unlike single-person motion generation~\cite{dhariwal2021diffusion,rombach2022high}, interactions require simultaneously modeling (i) long-range temporal causality within each individual and (ii) tightly coupled, dynamically evolving coordination across individuals~\cite{liang2024intergen,xu2024inter}. Small temporal or coordination errors quickly compound over time, leading to orientation inconsistencies, missed physical contacts, and interaction drift.

Recent diffusion models achieve high-fidelity motion synthesis~\cite{dhariwal2021diffusion,rombach2022high,dabral2023mofusion,zhang2022motiondiffuse} and have been extended to interactions~\cite{liang2024intergen,ponce2024in2in,cen2025ready,wu2025text2interact}. However, most interaction diffusion models denoise full sequences with bidirectional attention, implicitly relying on future context. This obscures causal structure, complicates streaming generation, and limits control over coordination roles (\eg, reactions or leader--follower behaviors). Autoregressive methods instead enforce causality~\cite{zhao2024dartcontrol,xiao2025motionstreamer,ruiz2025interact2ar,liu2026hint}, but often suffer from drift and boundary artifacts in long sequences, especially in interactions, where coordination errors compound across partners.

\begin{figure*}[t]
    \centering
    \includegraphics[width=1\linewidth]{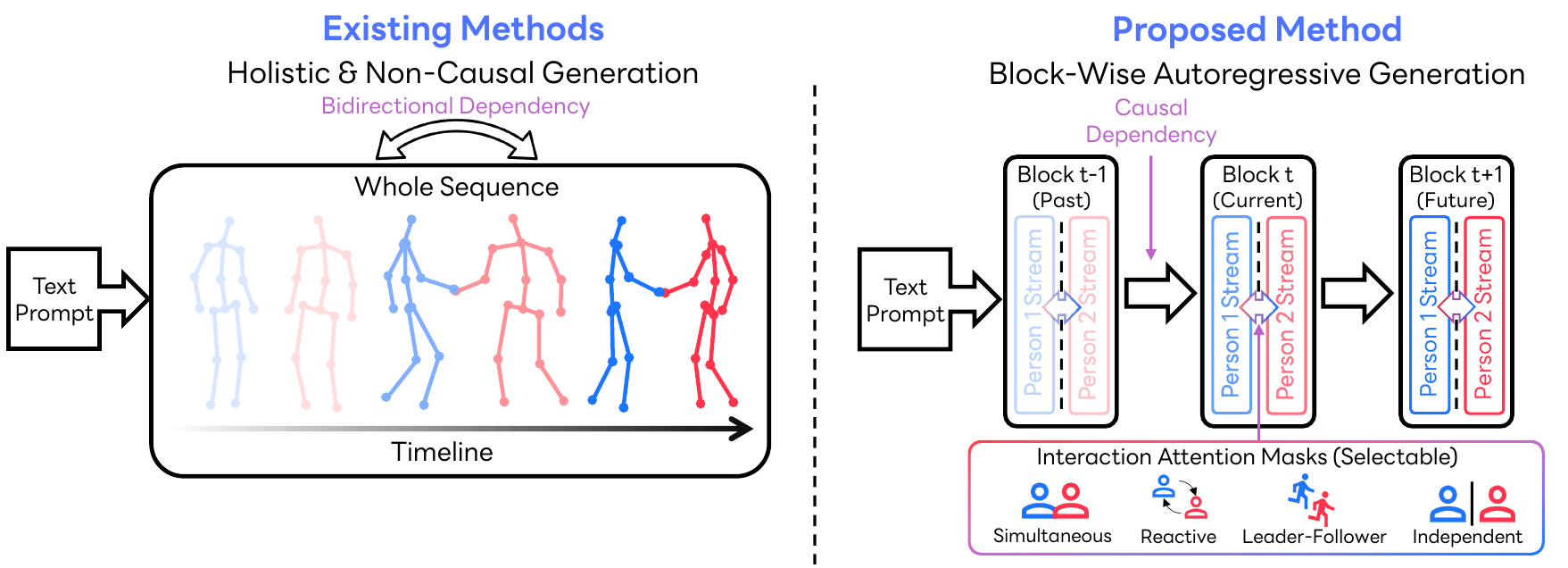}
    \caption{Comparison of existing interaction generation methods and InterCMDM\@. Left: prior methods perform full-sequence, non-causal generation. Right: InterCMDM uses \textit{block-wise} autoregressive generation with dual causal streams and interaction masks.}
    \label{fig:setting}
\end{figure*}

In this paper, we introduce \textbf{InterCMDM}, a \emph{block-causal latent diffusion} framework that combines diffusion-level realism with explicit causal interaction structure (\Fref{fig:setting}). InterCMDM generates motion block by block in the latent space of a temporal motion VAE, enabling long-horizon rollout.

InterCMDM is built on three core components. First, we propose the \textbf{Dual-Stream Causal Diffusion Transformer (DS-Causal-DiT)}, which models interaction as two coupled but structurally distinct temporal processes. Unlike prior methods that either concatenate both people into a single stream~\cite{ghosh2025duetgen,javed2025intermask} or handle them with separate attention modules~\cite{liang2024intergen,xu2024inter}, our design maintains separate causal streams while unifying interaction through a dual-stream attention mechanism. Causal self-attention preserves temporal ordering within each person, while cross-stream attention captures dynamic coordination, enabling joint modeling of intra-person causality and inter-person interaction.

Second, we propose \textbf{multi-task block attention masking}, a unified training strategy that exposes the model to diverse coordination structures, including simultaneous, reactive, leader--follower, and independent interactions. Each masking pattern defines distinct causal and cross-stream attention constraints. By randomly sampling these masks during training, the model learns generalized interaction dynamics. At inference time, selecting a specific mask enables controllable generation tailored to the desired coordination behavior.

Third, we reformulate diffusion for interaction generation through a \textbf{block-wise diffusion objective}. Rather than denoising entire sequences~\cite{ho2020denoising}, our model progressively denoises temporal blocks in latent space. Inspired by Diffusion Forcing for next-token prediction and single-person motion generation~\cite{chen2024diffusion,yu2026causal}, this formulation preserves causal dependencies across blocks, \ie, segments of frames, while enabling parallel denoising within each block. The resulting block-causal design supports stable long-horizon rollouts while mitigating drift.

Experiments on InterHuman and Inter-X show that InterCMDM achieves state-of-the-art text-motion alignment and realism. In long-horizon evaluations, latent rollout demonstrates good continuity while maintaining competitive local motion quality. Ablation studies further validate the benefits of dual-stream causal modeling and multi-task masking.

Our contributions are summarized as follows:
\begin{itemize}[leftmargin=*, topsep=0pt, itemsep=0pt, parsep=0pt]
  \item \textbf{DS-Causal-DiT:} a dual-stream diffusion transformer that enforces per-person temporal causality and models structured inter-person dependencies via dual-stream masked attention.
  \item \textbf{Multi-task block attention masking:} a unified framework for learning diverse coordination patterns and enabling controllable interaction generation.
  \item \textbf{Block-wise diffusion objective:} an autoregressive latent diffusion formulation enabling stable long-horizon interaction rollout.
  \item \textbf{State-of-the-art results} on InterHuman and Inter-X with improved realism, coherence, and interaction stability.
\end{itemize}

\section{Related Work}

\paragraph{Generative Human Motion Modeling.}
Large-scale datasets such as AMASS~\cite{mahmood2019amass}, BABEL~\cite{punnakkal2021babel}, and KIT-ML~\cite{plappert2016kit} have driven rapid progress in generative human motion modeling. Early RNN- and Transformer-based approaches~\cite{yan2019convolutional,zhao2020bayesian,guo2020action2motion,petrovich2021action,petrovich2022temos,athanasiou2022teach} struggled with long-range dependencies and semantic alignment. Diffusion models~\cite{ho2020denoising,dhariwal2021diffusion,rombach2022high} now dominate due to superior realism and diversity. Motion-space diffusion methods such as MDM~\cite{mdm2022human} and MotionDiffuse~\cite{zhang2022motiondiffuse} directly denoise trajectories, while latent variants including MLD~\cite{chen2023executing}, MotionLCM~\cite{dai2024motionlcm}, and SALAD~\cite{hong2025salad} improve efficiency. Recent extensions enhance controllability via retrieval augmentation~\cite{zhang2023remodiffuse,yu2025remogpt,li2025remomask}, guidance~\cite{karunratanakul2023guided,xie2024omnicontrol,liu2024programmable, watanabe2026projflow}, and compositional generation~\cite{barquero2024seamless,petrovich2024multi,zhang2025energymogen}. However, most diffusion models rely on full-sequence bidirectional denoising, implicitly assuming future context. This violates temporal causality and limits streaming and stable long-horizon generation, especially in human-to-human interaction settings.

\paragraph{Generative Human Interaction Modeling.}
Interaction modeling requires reasoning over both intra-person dynamics and inter-person coordination. ComMDM~\cite{shafir2024human} connects pretrained single-person models, while RIG~\cite{tanaka2023interaction}, InterGen~\cite{liang2024intergen} and Inter-X~\cite{xu2024inter} introduce interaction diffusion architectures and datasets. TIMotion~\cite{wang2025timotion} introduces temporal modeling and interaction mixing during the diffusion process. Recent work explores masked and autoregressive formulations. InterMask~\cite{javed2025intermask} adopts bidirectional masked modeling, and Interact2Ar~\cite{ruiz2025interact2ar} introduces autoregressive diffusion, but operates frame-by-frame in motion space, leading to drift and computational overhead. HINT~\cite{liu2026hint} improves realism, but relies on ground-truth prefix conditioning. Overall, existing interaction models either rely on full-sequence bidirectional denoising without explicit causality or perform fine-grained autoregression with accumulated drift. Neither approach simultaneously preserves diffusion fidelity and structured causal modeling.

\paragraph{Autoregressive Motion Generation.}
Autoregressive (AR) motion models enforce temporal causality by predicting future frames from past context. Discrete-token methods such as T2M-GPT~\cite{zhang2023t2m} and MotionGPT~\cite{jiang2023motiongpt} suffer from exposure bias~\cite{ning2023elucidating,schmidt2019generalization}. VQ-based approaches~\cite{guo2024momask,pinyoanuntapong2024mmm,zou2024parco} mitigate some limitations, but introduce quantization artifacts. Recent autoregressive diffusion models bridge AR and diffusion. Dart~\cite{zhao2024dartcontrol}, MARDM~\cite{meng2024rethinking}, and MotionStreamer~\cite{xiao2025motionstreamer} incorporate diffusion into causal frameworks. CMDM~\cite{yu2026causal} introduces causal diffusion forcing~\cite{chen2024diffusion} for autoregressive motion generation with token-wise causal attention and a temporal VAE for single-person settings.

However, extending these designs to interaction introduces additional challenges: structured cross-person dependencies, diverse coordination patterns, and amplified drift when two people must remain aligned. Our work builds upon CMDM's causal diffusion and temporal VAE framework, extending it from single-person motion generation to interaction modeling through: (1) a dual-stream architecture with unified masked attention for cross-person dependencies; (2) multi-task block attention masking for diverse coordination patterns; and (3) inference-time controllability via mask selection. These extensions enable structured two-person interaction generation beyond CMDM's single-stream design.

\section{Method}
\label{sec:method}
To enable structured causal two-person interaction generation, we propose \textbf{InterCMDM}, a block-causal latent diffusion framework for autoregressive interaction modeling. Unlike prior causal diffusion models designed for single-person motion, InterCMDM explicitly captures structured cross-person dependencies while preserving per-person temporal causality. The framework, as shown in~\Fref{fig:pipeline}, introduces three key components: (i) a \textbf{Dual-Stream Causal Diffusion Transformer (DS-Causal-DiT)} that factorizes interaction into separate causal streams and couples them through unified cross-stream attention; (ii) \textbf{multi-task block attention masking}, which enables controllable coordination behaviors within a single model; and (iii) a \textbf{block-wise diffusion objective} that supports long-horizon rollout in latent space. A temporal motion VAE provides a compact and efficient latent representation for block-wise generation.

\subsection{Problem Formulation}
Let $\mathbf{x}^{(1)}, \mathbf{x}^{(2)} \in \mathbb{R}^{T \times D}$ denote the motion sequences of two interacting people over $T$ frames, where $D$ is the feature dimension per frame (\eg, joint positions, rotations, and velocities), and let $\mathbf{c}$ be a text prompt. Our goal is to autoregressively generate $(\mathbf{x}^{(1)}, \mathbf{x}^{(2)})$ such that the motions are temporally coherent, physically plausible, and semantically aligned with $\mathbf{c}$.

We train a diffusion model in latent space, partitioning the latent sequences into non-overlapping temporal blocks of size $B$ and imposing a \textbf{block-causal} constraint: tokens may attend to tokens within the current block and all previous blocks, but never to future blocks. This design preserves diffusion parallelism within each block while enforcing an autoregressive structure across blocks.

\subsection{Temporal Motion VAE}

We encode each person independently into latent sequences using a shared temporal motion VAE\@. Both the encoder and decoder are built with 1D causal convolutions, as in CMDM~\cite{yu2026causal}, ensuring temporal causality during both encoding and decoding:
\begin{equation}
\mathbf{z}_\tau^{(i)} = \mathcal{E}_\phi(\mathbf{x}_{\le \tau}^{(i)}), \quad
\hat{\mathbf{x}}_\tau^{(i)} = \mathcal{D}_\psi(\mathbf{z}_{\le \tau}^{(i)}), 
\quad i \in \{1,2\}.
\end{equation}

The encoder temporally downsamples the sequence ($T' = T/4$), producing compact latent tokens of dimension $d_z$, each aligned with a local 4-frame motion window. This design enables efficient diffusion in latent space and avoids repeated decode--encode cycles during autoregressive rollout.

The VAE is trained with a variational objective that combines a reconstruction loss and a Kullback--Leibler divergence term for each person:
$\mathcal{L}_{\text{VAE}}=\sum_{i=1}^{2} \left(\mathcal{L}_{\text{rec}}^{(i)} + \beta D_{\text{KL}}^{(i)}\right)$.
This per-person formulation preserves separate latent representations, while the shared encoder--decoder parameters ensure consistent encoding across both individuals. During inference, motion can be decoded sequentially in real time, enabling streaming generation without access to future frames.

\begin{figure*}[t]
    \centering
    \includegraphics[width=1\linewidth]{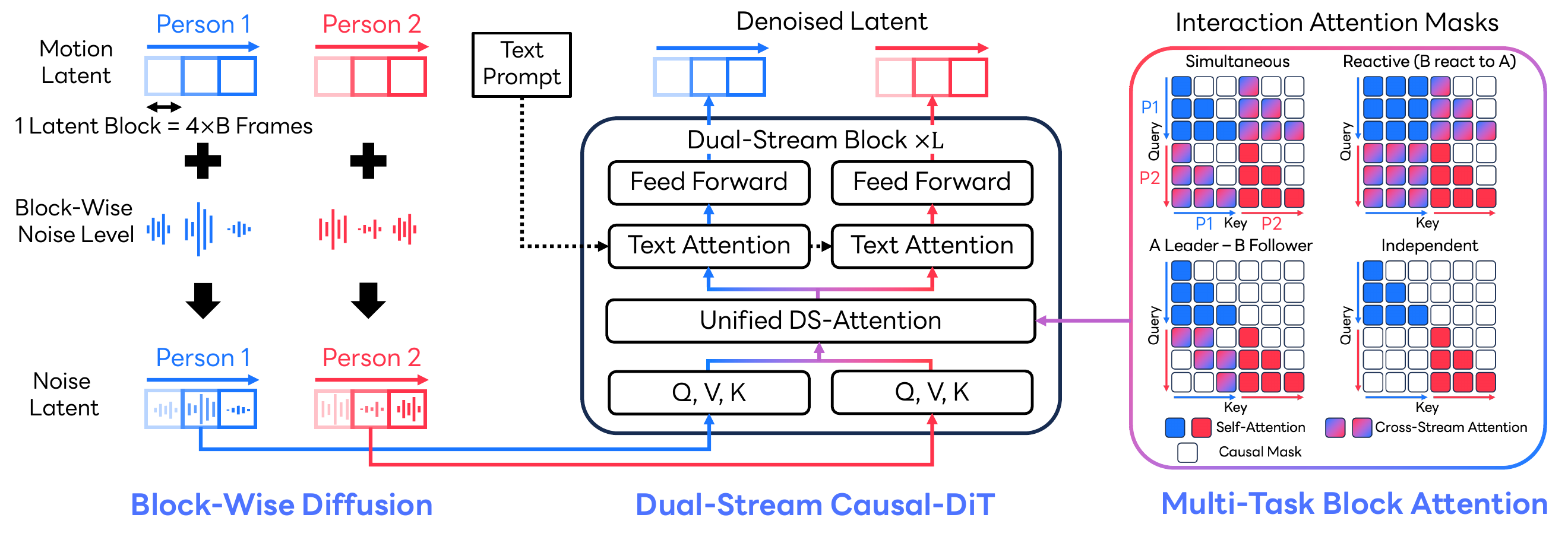}
    \caption{InterCMDM pipeline. Motion sequences are encoded into latent blocks and diffused via block-wise diffusion (left). The DS-Causal-DiT (center) processes two causal streams with unified dual-stream attention and text cross-attention. Multi-task block attention masks (right) impose structured causal and cross-stream constraints.}
    \label{fig:pipeline}
\end{figure*}

\subsection{Dual-Stream Causal Diffusion Transformer (DS-Causal-DiT)}

InterCMDM adopts a DS-Causal-DiT to explicitly model two-person interaction under block-wise diffusion. As shown in~\Fref{fig:pipeline}, motion latents from Person~1 and Person~2 are processed as two parallel streams within each temporal block, while block-wise noise levels are injected during denoising.

Unlike prior interaction diffusion models that either concatenate both people into a single stream~\cite{ghosh2025duetgen,javed2025intermask} or separate self- and cross-attention modules~\cite{liang2024intergen,xu2024inter}, our design maintains two structured causal streams and unifies their interaction within a single attention operation inside each dual-stream block.

\paragraph{Dual-Stream Block.}
Given noisy latent blocks $\mathbf{z}^{(1)}, \mathbf{z}^{(2)} \in \mathbb{R}^{T' \times d_z}$ for Person~1 and Person~2, DS-Causal-DiT applies $L$ stacked Dual-Stream blocks. Each block preserves separate temporal representations while enabling interaction modeling.

Let $\mathbf{h}_i^{(0)} = \mathrm{Embed}(\mathbf{z}^{(i)})$, $i \in \{1,2\}$. At layer $\ell$:
\begin{align}
[\tilde{\mathbf{h}}_1^{(\ell)}; \tilde{\mathbf{h}}_2^{(\ell)}]
&= \mathrm{DualStreamAttn}([\mathbf{h}_1^{(\ell-1)}; \mathbf{h}_2^{(\ell-1)}], \mathcal{M})
+ [\mathbf{h}_1^{(\ell-1)}; \mathbf{h}_2^{(\ell-1)}], \\
\hat{\mathbf{h}}_i^{(\ell)}
&= \mathrm{TextAttn}(\tilde{\mathbf{h}}_i^{(\ell)}, \mathbf{c}) + \tilde{\mathbf{h}}_i^{(\ell)}, \\
\mathbf{h}_i^{(\ell)}
&= \mathrm{FFN}(\hat{\mathbf{h}}_i^{(\ell)}) + \hat{\mathbf{h}}_i^{(\ell)}.
\end{align}
Details of other components, including the text encoder, positional encoding, and diffusion timestep embedding, are provided in the supplementary material.

\paragraph{Unified Dual-Stream Attention with Multi-Task Block Masking.}
\label{sec:multi_task_masking}
The core of DS-Causal-DiT is a unified masked attention mechanism that preserves per-person temporal causality while modeling structured cross-person dependencies via the attention mask $\mathcal{M}$. Within each block, separate query, key, and value projections are computed for Person~1 and Person~2, and the two streams are concatenated along the sequence dimension. A single masked attention operation is then applied to the concatenated sequence as shown in~\Fref{fig:pipeline}.

The unified dual-stream attention is controlled by a structured mask $\mathcal{M}$ that jointly encodes temporal causality and interaction patterns. Formally, $\mathcal{M}[i,j]$ determines whether token $i$ (from either stream) is allowed to attend to token $j$ in the concatenated sequence. The mask enforces block-causal constraints (restricting attention to the current block and all preceding blocks) and cross-stream interaction structures (specifying how the two people attend to each other). Different mask configurations enable distinct coordination modes.

\paragraph{Multi-Task Block Masking.}
Human interactions exhibit diverse coordination structures that differ in information flow and temporal dependency. Rather than training separate models, we adopt multi-task block masking to expose the model to multiple interaction structures during training. We define four mask types to cover representative and complementary coordination patterns:

\begin{enumerate}[leftmargin=*, topsep=0pt, itemsep=0pt, parsep=0pt]
    \item \textbf{Simultaneous:} Both people attend causally to their own past and to the other’s past.
    \item \textbf{Reactive:} One person conditions on the other’s completed motion before responding.
    \item \textbf{Leader--Follower:} The follower attends to the leader with a temporal lag.
    \item \textbf{Independent:} Cross-stream attention is disabled.
\end{enumerate}

These masks are designed to span symmetric, asymmetric, delayed, and decoupled interaction regimes, providing structured supervision over different cross-person dependency patterns. During training, mask types are randomly sampled to encourage generalization across coordination dynamics. 
At inference, selecting a mask directly determines the generated interaction behavior without retraining. The detailed formulation is provided in the supplementary material.

\subsection{Block-Wise Diffusion Objective}
We extend causal diffusion forcing~\cite{chen2024diffusion, yu2026causal} from frame-level noise to a \textbf{block-level} formulation in latent space. Unlike conventional diffusion models that apply a single noise level to the entire sequence, we assign an independent diffusion timestep to each temporal block. For clean latents $\mathbf{z}_0^{(1)}, \mathbf{z}_0^{(2)} \in \mathbb{R}^{T' \times d_z}$ of Person~1 and Person~2, we reformulate diffusion as a \textbf{block-causal latent process}. Each latent sequence is partitioned into $N_B = T'/B$ non-overlapping temporal blocks:
\begin{equation}
\mathbf{z}_0^{(i)} 
= 
\left[
\mathbf{z}_0^{(i,1)}, 
\mathbf{z}_0^{(i,2)}, 
\dots, 
\mathbf{z}_0^{(i,N_B)}
\right],
\quad i \in \{1,2\},
\end{equation}
where $\mathbf{z}_0^{(i,b)} \in \mathbb{R}^{B \times d_z}$ denotes the $b$-th block of person $i$.

\paragraph{Block-Level Diffusion Process.}
For each block $b$, we sample an independent diffusion timestep $k_b \in [0,K]$ and define the noisy latent block as:
\begin{equation}
\mathbf{z}_{k_b}^{(i,b)}
=
\sqrt{\bar{\alpha}_{k_b}} \, \mathbf{z}_0^{(i,b)}
+
\sqrt{1-\bar{\alpha}_{k_b}} \, \boldsymbol{\epsilon}^{(i,b)},
\quad
\boldsymbol{\epsilon}^{(i,b)} \sim \mathcal{N}(0,\mathbf{I}),
\end{equation}
where $\sqrt{\bar{\alpha}_{k_b}}$ and $\sqrt{1-\bar{\alpha}_{k_b}}$ are determined by the diffusion noise schedule.

As a result, the noisy latent sequence for person $i$ is:
$
\mathbf{z}_{\mathbf{k}}^{(i)}
=
\left[
\mathbf{z}_{k_1}^{(i,1)},
\dots,
\mathbf{z}_{k_{N_B}}^{(i,N_B)}
\right],
$
where $\mathbf{k} = (k_1, \dots, k_{N_B})$.

\paragraph{Block-Causal Denoising.}
The DS-Causal-DiT jointly predicts the noise residuals for both streams under the unified dual-stream attention mechanism. For example, under strict block-causal attention (\ie, the simultaneous mask), the prediction for block $b$ conditions only on the noisy latent blocks up to $b$:
\begin{equation}
\mathcal{L}_{\text{BW}}
=
\mathbb{E}_{\mathbf{k}, \boldsymbol{\epsilon}}
\left[
\sum_{b=1}^{N_B}
\sum_{i=1}^{2}
\left\|
\boldsymbol{\epsilon}^{(i,b)}
-
\boldsymbol{\mu}_\theta^{(i)}
\big(
\mathbf{z}_{\mathbf{k},\le b}^{(1)},
\mathbf{z}_{\mathbf{k},\le b}^{(2)},
k_b,
\mathbf{c}
\big)
\right\|_2^2
\right],
\end{equation}
where $\mathbf{z}_{\mathbf{k},\le b}^{(i)}$ denotes the noisy latent blocks of person $i$ from block $1$ to $b$. The denoiser outputs a pair of residual predictions, $\boldsymbol{\mu}_\theta = (\boldsymbol{\mu}_\theta^{(1)}, \boldsymbol{\mu}_\theta^{(2)})$, corresponding to Person 1 and Person 2, respectively.

This block-wise formulation preserves diffusion parallelism within each block while enforcing causal dependencies across blocks, enabling stable autoregressive rollout without full-sequence denoising.

\paragraph{Discussion.}
Unlike full-sequence diffusion
$
\mathbf{z}_k
=
\sqrt{\bar{\alpha}_k}\mathbf{z}
+
\sqrt{1-\bar{\alpha}_k}\boldsymbol{\epsilon},
$
our block-wise formulation applies noise at the block level and enforces temporal causality through block-restricted attention. This design bridges diffusion modeling and autoregressive generation by retaining parallel intra-block denoising while supporting causal rollout across blocks.

\paragraph{Autoregressive Inference (Latent Rollout).}
At inference, generation proceeds autoregressively over latent blocks. Some existing methods~\cite{ruiz2025interact2ar,liu2026hint} use a ground-truth prefix to initialize generation. In contrast, our method initializes the first block ($b=1$) from pure Gaussian noise, $\mathbf{z}_K^{(1,1)}, \mathbf{z}_K^{(2,1)} \sim \mathcal{N}(0, \mathbf{I})$, and denoises it over $K$ diffusion steps using the selected attention mask $\mathcal{M}$. For subsequent blocks $b > 1$, we sample:
\begin{equation}
\mathbf{z}_0^{(1,b)}, \mathbf{z}_0^{(2,b)}
\sim
p_\theta
\big(
\cdot \mid
\mathbf{z}_0^{(1,<b)},
\mathbf{z}_0^{(2,<b)},
\mathbf{c}
\big),
\end{equation}
where $\mathbf{z}_0^{(i,<b)}$ denotes the previously generated clean latent blocks of person $i$. The motion sequence is then recovered through the temporal motion VAE\@. This latent rollout avoids repeated decode--encode cycles for prefix construction, mitigating cumulative reconstruction drift and improving boundary continuity during long-horizon synthesis.
\section{Experiments}
\label{sec:experiments}

\subsection{Experimental Setup}

\paragraph{Datasets.}
We evaluate InterCMDM on two text-to-interaction benchmarks: InterHuman~\cite{liang2024intergen} and Inter-X~\cite{xu2024inter}.  
InterHuman contains 7,779 sequences paired with 23,337 captions and is represented using the 22-joint AMASS skeleton (262-dimensional features per frame, including global joint positions, velocities, and rotations).  
Inter-X comprises 11,388 sequences with three annotations per sequence and adopts the 56-joint SMPL-X representation (336-dimensional features) to capture detailed hand and facial motion.  
Both datasets are recorded at 30 fps, and we follow the standard train/validation/test splits.

\paragraph{Evaluation Metrics.}
Following prior work~\cite{liang2024intergen,javed2025intermask,xu2024inter}, we report R-Precision (Top-1/2/3), Fréchet Inception Distance (FID), Multi-Modal Distance (MM Dist), Diversity, and Multi-Modality.  
Lower FID and MM Dist indicate better realism and alignment, while higher R-Precision and Multi-Modality reflect stronger semantic consistency and diversity.

\paragraph{Implementation Details.}
The Temporal Motion VAE downsamples motion by 4$\times$, producing 64-dimensional latents for every 4 frames. For VAE training, motion sequences are segmented into 64-frame clips with a sliding window of 10 frames. The VAE is trained for 50 epochs using AdamW with a learning rate of $2{\times}10^{-4}$, a batch size of 256, and a learning-rate schedule that decays by 0.1 twice. Gradient clipping with a maximum norm of 1.0 is applied to stabilize training. VAE training takes approximately 1 hour on a single NVIDIA A100 GPU.

DS-Causal-DiT consists of 8 transformer layers with 4 attention heads and a hidden size of 512, using block-causal attention with block size $B=3$. Text conditioning is provided by a frozen DistilBERT encoder. For DS-Causal-DiT training, motion sequences from InterHuman and Inter-X are clipped or padded to fixed lengths of 300 and 156 frames, respectively. The streams of Person~1 and Person~2 are randomly shuffled during training.

Attention mask types are randomly sampled with probabilities of 60\% for Simultaneous, 20\% for Reactive, 10\% for Leader--Follower, and 10\% for Independent. This weighting reflects the empirical distribution of coordination patterns in our interaction datasets, where simultaneous interactions are most common, while reactive, leader--follower, and independent movements occur less frequently. It prioritizes the dominant interaction pattern while maintaining exposure to diverse coordination behaviors.

We train DS-Causal-DiT for 500 epochs using AdamW with a learning rate of $10^{-4}$ and a batch size of 64. We use Flow Matching~\cite{lipman2022flow, ma2024sit, albergo2023stochastic, albergo2022building} as the diffusion scheduler. Training takes approximately 8 hours on a single NVIDIA A100 GPU. During inference, we use 50 diffusion steps, a classifier-free guidance scale of 3.0 for InterHuman and 1.5 for Inter-X. We follow the block-wise sampling schedule of~\cite{yu2026causal}.

\subsection{Quantitative Results}
\paragraph{Results on InterHuman.}
We compare InterCMDM with state-of-the-art interaction generation methods across three paradigms:
(1) \emph{Recent diffusion-based} (InterGen~\cite{liang2024intergen}, MoMat-MoGen~\cite{cai2024digital}, in2IN~\cite{ponce2024in2in}, TIMotion~\cite{wang2025timotion}, InterActor~\cite{wu2025text2interact});
(2) \emph{Generative masked modeling} (InterMask~\cite{javed2025intermask});
(3) \emph{Autoregressive} (Interact2Ar~\cite{ruiz2025interact2ar}, HINT~\cite{liu2026hint}).

\begin{table}[t]
\caption{Results on the InterHuman test set. Methods marked with $\dagger$ are autoregressive approaches that use several initial ground-truth frames during evaluation. The average is reported over 10 runs with 95\% confidence intervals. For all tables, \textbf{bold} indicates the best result, and \underline{underline} denotes the second-best result.}
\label{tab:interhuman_results}
\centering
\resizebox{\linewidth}{!}{
\begin{tabular}{l|ccc|cc|cc}
\toprule
\multirow{2}{*}{Method} 
& \multicolumn{3}{c|}{R-Precision$\uparrow$} 
& \multirow{2}{*}{FID$\downarrow$} 
& \multirow{2}{*}{MM Dist$\downarrow$} 
& \multirow{2}{*}{Diversity$\rightarrow$}
& \multirow{2}{*}{MModality$\uparrow$} \\
\cline{2-4}
& Top-1 & Top-2 & Top-3 & & & & \\
\midrule
Real Motions 
& $0.452^{\pm .008}$ 
& $0.610^{\pm .009}$ 
& $0.701^{\pm .008}$ 
& $0.273^{\pm .007}$ 
& $3.755^{\pm .008}$ 
& $7.948^{\pm .064}$ 
& -- \\
\midrule
InterGen 
& $0.371^{\pm .010}$ 
& $0.515^{\pm .012}$ 
& $0.624^{\pm .010}$ 
& $5.918^{\pm .079}$ 
& $5.108^{\pm .014}$ 
& $7.387^{\pm .029}$ 
& $\underline{2.141^{\pm .063}}$ \\
MoMat-MoGen 
& $0.449^{\pm .004}$ 
& $0.591^{\pm .003}$ 
& $0.666^{\pm .004}$ 
& $5.674^{\pm .085}$ 
& $3.790^{\pm .001}$ 
& $8.021^{\pm .350}$ 
& $1.295^{\pm .023}$ \\
in2IN 
& $0.425^{\pm .008}$ 
& $0.576^{\pm .008}$ 
& $0.662^{\pm .009}$ 
& $5.535^{\pm .120}$ 
& $3.803^{\pm .002}$ 
& $\underline{7.953^{\pm .047}}$ 
& $1.215^{\pm .023}$ \\
InterMask 
& $0.449^{\pm .004}$ 
& $0.599^{\pm .005}$ 
& $0.683^{\pm .004}$ 
& $5.154^{\pm .061}$ 
& $3.790^{\pm .002}$ 
& $\mathbf{7.944^{\pm .033}}$ 
& $1.737^{\pm .020}$ \\
TIMotion 
& $\underline{0.501^{\pm .005}}$ 
& $\underline{0.656^{\pm .006}}$ 
& $\underline{0.734^{\pm .006}}$ 
& ${4.702^{\pm .069}}$ 
& $\underline{3.769^{\pm .021}}$ 
& $7.943^{\pm .034}$ 
& $1.005^{\pm .020}$ \\
InterActor 
& ${0.483^{\pm .005}}$ 
& ${0.638^{\pm .005}}$ 
& ${0.717^{\pm .005}}$ 
& $5.191^{\pm .055}$ 
& ${3.778^{\pm .001}}$ 
& $7.900^{\pm .030}$ 
& $1.051^{\pm .031}$ \\
HINT$^\dagger$
& - 
& -
& $0.672^{\pm .004}$ 
& $\mathbf{3.100^{\pm .035}}$ 
& $3.796^{\pm .001}$ 
& $7.898^{\pm .023}$ 
& - \\
\midrule
\textbf{InterCMDM} 
& $\mathbf{0.529^{\pm .007}}$ 
& $\mathbf{0.678^{\pm .008}}$ 
& $\mathbf{0.757^{\pm .005}}$ 
& $\underline{4.492^{\pm .087}}$ 
& $\mathbf{3.765^{\pm .002}}$ 
& $8.065^{\pm .034}$ 
& $\mathbf{3.044^{\pm 0.104}}$ \\
\bottomrule
\end{tabular}
}
\end{table}

As shown in Table~\ref{tab:interhuman_results}, InterCMDM achieves the strongest overall performance on InterHuman among fully generative, representation-consistent methods.
Compared to the strongest diffusion baseline TIMotion~\cite{wang2025timotion}, InterCMDM improves R-Precision from 0.501 to 0.529 (+0.028) on Top-1, indicating stronger text–motion alignment.
InterCMDM also achieves the lowest MM Dist (3.765), demonstrating improved semantic consistency.
While HINT$^\dagger$ reports a lower FID (3.100), it relies on autoregressive generation with initial ground-truth frames; among fully generative diffusion-based methods, InterCMDM achieves the best FID (4.492).
InterCMDM also achieves the highest MModality and maintains diversity close to real motions, indicating improved alignment and realism without sacrificing variation.

\begin{table}[t]
\caption{Results on the Inter-X test set. Methods marked with $\dagger$ are autoregressive approaches that use several initial ground-truth frames during evaluation. Methods marked with $*$ are trained using a different motion representation.}
\label{tab:interx_results}
\centering
\resizebox{\linewidth}{!}{
\begin{tabular}{l|ccc|cc|cc}
\toprule
\multirow{2}{*}{Method} 
& \multicolumn{3}{c|}{R-Precision$\uparrow$} 
& \multirow{2}{*}{FID$\downarrow$} 
& \multirow{2}{*}{MM Dist$\downarrow$} 
& \multirow{2}{*}{Diversity$\rightarrow$}
& \multirow{2}{*}{MModality$\uparrow$} \\
\cline{2-4}
& Top-1 & Top-2 & Top-3 & & & & \\
\midrule
Real Motions 
& $0.429^{\pm .004}$ 
& $0.626^{\pm .003}$ 
& $0.736^{\pm .003}$ 
& $0.002^{\pm .000}$ 
& $3.536^{\pm .013}$ 
& $9.734^{\pm .078}$ 
& -- \\
\midrule
InterGen 
& $0.207^{\pm .004}$ 
& $0.335^{\pm .005}$ 
& $0.429^{\pm .005}$ 
& $5.207^{\pm .216}$ 
& $9.580^{\pm .011}$ 
& $7.788^{\pm .208}$ 
& $\mathbf{3.686^{\pm .052}}$ \\
InterMask 
& $0.403^{\pm .005}$ 
& $0.595^{\pm .004}$ 
& $0.705^{\pm .005}$ 
& $0.399^{\pm .013}$ 
& $3.705^{\pm .017}$ 
& $9.046^{\pm .073}$ 
& $2.261^{\pm .081}$ \\
TIMotion 
& $0.411^{\pm .005}$ 
& $0.597^{\pm .006}$ 
& $0.707^{\pm .004}$ 
& $0.261^{\pm .014}$ 
& $3.737^{\pm .015}$ 
& $9.112^{\pm .079}$ 
& $2.475^{\pm .075}$ \\
Interact2Ar$^{*\dagger}$
& $\underline{0.441^{\pm .00?}}$ 
& $\underline{0.631^{\pm .00?}}$ 
& $\underline{0.737^{\pm .00?}}$ 
& $\mathbf{0.148^{\pm .01?}}$ 
& $\underline{3.581^{\pm .01?}}$ 
& $\underline{9.147^{\pm .06?}}$ 
& $1.529^{\pm .05?}$ \\
HINT$^\dagger$
& - 
& -
& $0.682^{\pm .003}$ 
& $0.278^{\pm .012}$ 
& $4.007^{\pm .016}$ 
& $8.886^{\pm .066}$ 
& - \\
\midrule
\textbf{InterCMDM} 
& $\mathbf{0.523^{\pm .009}}$ 
& $\mathbf{0.721^{\pm .008}}$ 
& $\mathbf{0.820^{\pm .007}}$ 
& $\underline{0.215^{\pm .014}}$ 
& $\mathbf{3.154^{\pm .021}}$ 
& $\mathbf{9.490^{\pm .120}}$ 
& $\underline{2.391^{\pm .136}}$  \\
\bottomrule
\end{tabular}
}
\end{table}
\begin{table}[t]
\caption{Long-horizon interaction generation on InterHuman. We evaluate per-segment quality and transition smoothness at segment boundaries. \textbf{B/NB Ratio}: boundary-to-non-boundary position discontinuity ratio (1.0 = perfect continuity).}
\label{tab:long_horizon}
\centering
\resizebox{\linewidth}{!}{
\begin{tabular}{ll|ccc|ccc}
\toprule
\multirow{2}{*}{Model} & \multirow{2}{*}{Strategy}
& \multicolumn{3}{c|}{Segment Quality} 
& \multicolumn{3}{c}{Transition Smoothness} \\
\cline{3-8}
& & FID$\downarrow$ & R-Top1$\uparrow$ & MM Dist$\downarrow$
& Pos Disc.$\downarrow$ & B/NB Ratio$\downarrow$ & Skating$\downarrow$ \\
\midrule
InterMask & AR Rollout
& $27.239^{\pm .39}$ & $0.320^{\pm .008}$ & $3.864^{\pm .018}$
& $0.048^{\pm .002}$ & $2.51^{\pm .07}$ & $0.172^{\pm .004}$ \\
\midrule
\multirow{3}{*}{InterCMDM} 
& Naive
& $\mathbf{18.37^{\pm .37}}$ & $\mathbf{0.423^{\pm .008}}$ & $\mathbf{3.775^{\pm .011}}$
& $1.375^{\pm .053}$ & $98.58^{\pm 1.21}$ & $0.260^{\pm .006}$ \\
& Composition
& $25.758^{\pm .35}$ & $0.338^{\pm .008}$ & $3.823^{\pm .014}$
& $\underline{0.026^{\pm .001}}$ & $\underline{1.91^{\pm .05}}$ & $\mathbf{0.136^{\pm .003}}$ \\
& Latent Rollout
& $\underline{25.104^{\pm .40}}$ & $\underline{0.352^{\pm .007}}$ & $\underline{3.798^{\pm .013}}$
& $\mathbf{0.024^{\pm .001}}$ & $\mathbf{1.66^{\pm .04}}$ & $\underline{0.157^{\pm .004}}$ \\
\bottomrule
\end{tabular}
}
\end{table}
\paragraph{Results on Inter-X.}
We further evaluate InterCMDM on the more challenging Inter-X benchmark. Table~\ref{tab:interx_results} shows that InterCMDM achieves the strongest overall performance on Inter-X among methods trained under comparable settings.
Compared to InterMask, InterCMDM improves R-Precision from 0.403 to 0.523 (+0.120) on Top-1, from 0.595 to 0.721 (+0.126) on Top-2, and from 0.705 to 0.820 (+0.115) on Top-3.
InterCMDM also achieves the lowest MM Dist (3.154) and the best diversity score (9.490), indicating better semantic alignment and motion variation.
Although Interact2Ar$^{*\dagger}$ reports a lower FID (0.148), it is trained with a different motion representation and uses initial ground-truth frames during evaluation.
Under fully generative and representation-consistent settings, InterCMDM achieves the best overall balance between alignment, realism, and diversity.
These results show that block-causal diffusion and unified dual-stream attention effectively model fine-grained full-body interactions.

\paragraph{Long-Horizon Interaction Generation.}
To evaluate InterCMDM on long-horizon generation, we compare: (1) InterMask with an autoregressive adaptation that extends generation using the most recent latent; and (2) three autoregressive strategies with our model: (a) Naive, which concatenates independently generated segments; (b) Composition, which follows the standard approach with repeated decode--encode cycles (Generate → Decode → Prefix → Encode → Generate); and (c) Latent Rollout, enabled by block-wise diffusion (Generate → Latent → Generate). We generate 64 long-horizon sequences on InterHuman, each consisting of 32 caption--duration pairs ($>120$ frames each), and use 32 segments and 31 transitions per sequence for evaluation.

We evaluate long-horizon generation on InterHuman in terms of \textit{segment quality} and \textit{transition smoothness}. 
Segment quality is measured using FID, R-Precision (Top-1), and MM Distance with respect to real motion clips. For FID, we compute statistics from real InterHuman clips that share the same prompt as the evaluation caption.
Transition smoothness is assessed at segment boundaries using three metrics: 
\textbf{Position Discrepancy (Pos Disc.)} (mean per-joint position error between consecutive segments), 
\textbf{Boundary/Non-Boundary (B/NB) Ratio} (boundary-to-intra-segment discontinuity ratio; 1.0 indicates perfect continuity), 
and \textbf{Skating} (average horizontal foot velocity during ground contact).

Table~\ref{tab:long_horizon} shows that Naive generation achieves the best per-segment quality (FID 18.37, Top1 0.423) since segments are generated independently, but it suffers from severe boundary artifacts (B/NB 98.58). 
In contrast, Latent Rollout improves segment quality over InterMask AR Rollout (FID 25.10 vs.\ 27.24, Top1 0.352 vs.\ 0.320). 
Latent Rollout achieves the lowest position discontinuity (0.024) and B/NB ratio (1.66), with reduced skating compared with InterMask (0.157 vs.\ 0.172), indicating smoother transitions at segment boundaries. 
Overall, block-wise diffusion with latent rollout provides a practical balance between motion quality, temporal continuity, and streaming latency.

\begin{figure*}[t]
    \centering
    \includegraphics[width=1\linewidth]{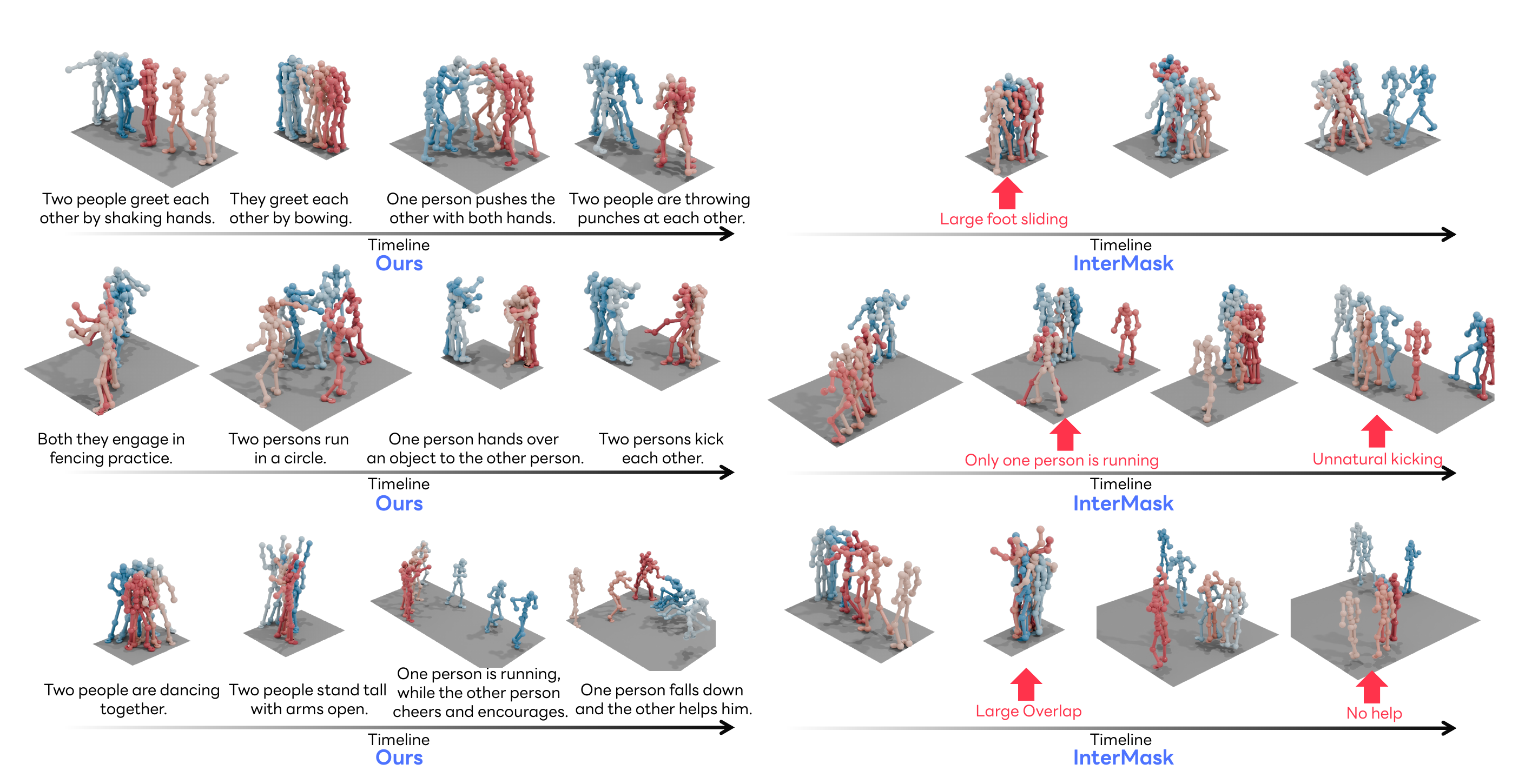}
    \caption{Qualitative comparison of long-horizon motion generation between InterCMDM and InterMask. Our results demonstrate more continuous and coherent interactions. For clarity, each long sequence is divided into shorter segments aligned with their corresponding captions. Please refer to the supplementary videos for full-length sequences.}
    \label{fig:qualitative}
\end{figure*}

\subsection{Qualitative Results}

\Fref{fig:qualitative} presents qualitative comparisons between InterCMDM and InterMask on diverse long-horizon interaction scenarios. As shown in the figure, InterCMDM generates coordinated and semantically consistent behaviors across a wide range of interactions, including greeting by shaking hands or bowing, fencing practice, kicking exchanges, cheering while the other runs, and helping a fallen partner.

Across these examples, InterCMDM preserves mutual facing, relative positioning, and action--reaction consistency over time. In contrast, InterMask exhibits noticeable artifacts, including large foot sliding, spatial overlap between two people, unnatural kicking motions, missing reaction behaviors (\eg, no helping action), and cases where only one person performs the intended action. These errors indicate weakened coordination and accumulated drift in long sequences.

Thanks to its continuous latent representation and block-causal structure, InterCMDM maintains smooth temporal transitions, stable contact patterns, and coherent inter-person alignment throughout extended timelines. Additional examples and video results are provided in the supplementary material.

\subsection{Analysis}

To investigate the impact of each component in InterCMDM, we conduct comprehensive ablation studies on the InterHuman dataset.

\begin{table*}[t]
\caption{
Ablation study of InterCMDM. We evaluate attention type (bidirectional vs. causal with block size $B$) and block design (single-stream vs. dual-stream).
}
\label{tab:ablation_architecture}
\centering
\small
\resizebox{\textwidth}{!}{
\begin{tabular}{l c | ccc | cc}
\toprule
Attention & Block Design 
& R-Top1$\uparrow$ 
& R-Top2$\uparrow$ 
& R-Top3$\uparrow$ 
& FID$\downarrow$ 
& MM Dist$\downarrow$ \\
\midrule
Bidirectional & Single
& $0.535^{\pm.003}$ 
& $0.692^{\pm.003}$ 
& $0.771^{\pm.002}$ 
& $\mathbf{4.456^{\pm.058}}$ 
& $3.758^{\pm.016}$ \\
Bidirectional & Hybrid
& $0.538^{\pm.003}$ 
& $0.693^{\pm.003}$ 
& $0.768^{\pm.002}$ 
& $4.468^{\pm.061}$ 
& $3.760^{\pm.017}$ \\
Bidirectional & Dual 
& $\mathbf{0.549^{\pm.003}}$ 
& $\mathbf{0.703^{\pm.002}}$ 
& $\mathbf{0.778^{\pm.002}}$ 
& $4.562^{\pm.064}$ 
& $\mathbf{3.754^{\pm.015}}$ \\

\midrule

Causal ($B=3$) & Single 
& $0.484^{\pm.003}$ 
& $0.641^{\pm.003}$ 
& $0.717^{\pm.003}$ 
& $5.068^{\pm.071}$ 
& $3.782^{\pm.018}$ \\

Causal ($B=3$) & Hybrid
& $0.499^{\pm.003}$ 
& $0.652^{\pm.003}$ 
& $0.734^{\pm.003}$ 
& $4.577^{\pm.063}$ 
& $3.774^{\pm.017}$ \\

Causal ($B=3$) & Dual 
& $0.529^{\pm.007}$ 
& $0.678^{\pm.008}$ 
& $0.757^{\pm.005}$ 
& $4.492^{\pm.087}$ 
& $3.765^{\pm.002}$ \\

\midrule

Causal ($B=1$) & Dual 
& $0.514^{\pm.007}$ 
& $0.663^{\pm.007}$ 
& $0.735^{\pm.009}$ 
& $4.551^{\pm.098}$ 
& $3.773^{\pm.002}$ \\

Causal ($B=5$) & Dual 
& $0.532^{\pm.007}$ 
& $0.684^{\pm.006}$ 
& $0.763^{\pm.004}$ 
& $4.545^{\pm.097}$ 
& $3.763^{\pm.002}$ \\
\bottomrule
\end{tabular}
}
\end{table*}

\paragraph{DS-Causal-DiT Architecture Validation.}
We validate DS-Causal-DiT on InterHuman by comparing dual-stream vs.\ single-stream designs and causal vs.\ bidirectional attention. Table~\ref{tab:ablation_architecture} shows three main findings. 

First, under causal attention ($B=3$), the dual-stream model significantly outperforms the single-stream variant (Top-1: 0.529 vs.\ 0.484; FID: 4.492 vs.\ 5.068). 
The hybrid design, \ie, using dual-stream blocks in the first half of the network and single-stream blocks in the second half, improves over pure single-stream but remains inferior to the full dual-stream architecture, indicating that consistent two-person modeling across all layers is important. 

Second, although bidirectional attention achieves a slightly higher Top-1 score (0.549) by accessing future blocks, it does not support autoregressive generation and yields worse transition smoothness in long-horizon generation. Among bidirectional models, dual-stream achieves a good balance of alignment and realism.

Third, block size influences performance. While $\#B=5$ achieves the highest alignment (Top-1: 0.532), it introduces additional latency in online generation due to larger temporal blocks. In contrast, $\#B=3$ yields the best FID (4.492) and enables lower-delay streaming generation. Token-level blocks ($\#B=1$) underperform in both alignment and realism.

Overall, separate causal streams with unified dual-stream attention and a moderate block size are essential for high-quality and real-time generation.

\begin{table*}[t]
\caption{
Comparison of model variants. We evaluate different stream designs and parameter-sharing strategies. In the dual-stream shared-weights variant, Person~1 and Person~2 share the same layer weights except for the unified attention layers.
}
\label{tab:ablation_parameters}
\centering
\small
\resizebox{\textwidth}{!}{
\begin{tabular}{l c | ccc | cc}
\toprule
Model Variants & \#Params
& R-Top1$\uparrow$ 
& R-Top2$\uparrow$ 
& R-Top3$\uparrow$ 
& FID$\downarrow$ 
& MM Dist$\downarrow$ \\

\midrule

Single-stream, shared weights & 39M
& $0.484^{\pm.003}$ 
& $0.641^{\pm.003}$ 
& $0.717^{\pm.003}$ 
& $5.068^{\pm.071}$ 
& $3.782^{\pm.018}$ \\

Dual-stream, shared weights& 47M
& $0.506^{\pm.007}$ 
& $0.658^{\pm.007}$ 
& $0.736^{\pm.008}$ 
& $4.672^{\pm.109}$ 
& $3.774^{\pm.02}$ \\

Dual-stream, separate weights (ours) & 77M
& $\mathbf{0.529^{\pm.007}}$ 
& $\mathbf{0.678^{\pm.008}}$ 
& $\mathbf{0.757^{\pm.005}}$ 
& $\mathbf{4.492^{\pm.087}}$ 
& $\mathbf{3.765^{\pm.002}}$ \\

\bottomrule
\end{tabular}
}
\end{table*}

\paragraph{Stream Parameter Sharing.}
We further analyze whether the gains come from maintaining two structured streams or from using person-specific parameters.
Table~\ref{tab:ablation_parameters} compares single-stream shared weights, dual-stream shared weights, and our dual-stream separate-weights design.
Using a dual-stream architecture with shared person weights improves over the single-stream baseline (Top-1: 0.506 vs.\ 0.484; FID: 4.672 vs.\ 5.068), showing that explicit stream separation provides the main structural benefit.
Allowing separate stream weights further improves performance (Top-1: 0.529; FID: 4.492), suggesting that person-specific parameters help capture role-asymmetric interactions such as leader/follower or actor/reactor behaviors.
This is especially useful in our causal setting, where the two streams may have different information availability under different masks.

\begin{table}[t]
\centering
\caption{Reaction generation on InterHuman. Person~1 is given as ground truth, and Person~2 is generated as the response.}
\label{tab:reaction_generation}
\resizebox{\columnwidth}{!}{%
\begin{tabular}{lc|ccc|cc|c}
\toprule
\multirow{2}{*}{Training Mask} 
& \multirow{2}{*}{Inference Mask} 
& \multicolumn{3}{c|}{R-Precision$\uparrow$} 
& \multirow{2}{*}{FID$\downarrow$} 
& \multirow{2}{*}{MM Dist$\downarrow$} 
& \multirow{2}{*}{Diversity$\rightarrow$} \\
\cline{3-5}
& & Top-1 & Top-2 & Top-3 & & & \\
\midrule
\multicolumn{2}{c|}{Real Motions}
& $0.432^{\pm .005}$ & $0.607^{\pm .004}$ & $0.708^{\pm .004}$
& $0.292^{\pm .009}$ & $3.786^{\pm .000}$ & $7.744^{\pm .040}$ \\ 
\multicolumn{2}{c|}{InterMask} 
& $0.427^{\pm .007}$ & $0.579^{\pm .006}$ & $0.674^{\pm .005}$ 
& $3.149^{\pm .060}$ & $3.799^{\pm .001}$ & $7.678^{\pm .041}$ \\ \midrule
\multicolumn{2}{c|}{InterCMDM} \\ 
Simultaneous & \multirow{3}{*}{Reactive} 
& $0.434^{\pm .006}$ & $0.597^{\pm .007}$ & $0.685^{\pm .006}$ 
& $2.585^{\pm .050}$ & $3.786^{\pm .001}$ & $7.686^{\pm .044}$ \\
Reactive & 
& $0.451^{\pm .006}$ & $0.612^{\pm .006}$ & $0.698^{\pm .006}$
& $2.216^{\pm .049}$ & $3.778^{\pm .002}$ & $7.843^{\pm .037}$ \\
Multi-task& 
& $\mathbf{0.484^{\pm .005}}$ & $\mathbf{0.642^{\pm .006}}$ & $\mathbf{0.727^{\pm .006}}$
& $\mathbf{2.121^{\pm .033}}$ & $\mathbf{3.771^{\pm .002}}$ & $\mathbf{7.751^{\pm .019}}$ \\
\bottomrule
\end{tabular}
}
\end{table}

\paragraph{Multi-Task Block Attention Masking.}
Our proposed multi-task block attention masking strategy enables a unified model to capture diverse interaction patterns through flexible attention control. We evaluate multi-task masking under a reaction generation setting, where Person~1 is given and Person~2 is generated as the response. Table~\ref{tab:reaction_generation} compares InterCMDM variants with InterMask.

Multi-task training achieves the best performance across all metrics. Compared to training with only the simultaneous mask (Top-1: 0.434; FID: 2.585), multi-task masking improves alignment (Top-1: 0.484) and reduces FID to 2.121. The reaction-only model also improves over the standard variant but remains inferior to multi-task training.

Compared to InterMask, InterCMDM with multi-task masking achieves higher alignment (0.484 vs.\ 0.427 Top-1) and substantially lower FID (2.121 vs.\ 3.149). These results show that multi-task block attention masking allows our model to adapt to other tasks such as reaction generation without dedicated training.

\begin{table}[t]
\centering
\caption{Comparison of attention mask types at training and inference. We evaluate InterCMDM using different attention mask configurations on InterHuman.}
\label{tab:attention_mask_comparison}
\resizebox{\linewidth}{!}{%
\begin{tabular}{cc|ccc|c|c|c}
\toprule
\multirow{2}{*}{Training Mask} 
& \multirow{2}{*}{Inference Mask} 
& \multicolumn{3}{c|}{R-Precision$\uparrow$} 
& \multirow{2}{*}{FID$\downarrow$} 
& \multirow{2}{*}{MM Dist$\downarrow$} 
& \multirow{2}{*}{Diversity$\rightarrow$} \\
\cline{3-5}
& & Top-1 & Top-2 & Top-3 & & & \\
\midrule
\multicolumn{2}{c|}{Real Motions}
& $0.452^{\pm .008}$ 
& $0.610^{\pm .009}$ 
& $0.701^{\pm .008}$ 
& $0.273^{\pm .007}$ 
& $3.755^{\pm .008}$ 
& $7.948^{\pm .064}$ 
\\
\midrule
Independent & Independent
& $0.492^{\pm .007}$ 
& $0.638^{\pm .005}$ 
& $0.727^{\pm .006}$ 
& $6.058^{\pm .130}$ 
& $3.782^{\pm .001}$ 
& $\mathbf{8.060^{\pm .039}}$
\\ \cline{1-2}
Simultaneous & \multirow{2}{*}{Simultaneous}  
& $0.498^{\pm .008}$ 
& $0.653^{\pm .008}$ 
& $0.730^{\pm .006}$ 
& $4.621^{\pm .055}$ 
& $3.775^{\pm .002}$ 
& $8.088^{\pm .032}$
\\
Multi-task (Even) & 
& $0.509^{\pm .007}$ 
& $0.664^{\pm .007}$ 
& $0.740^{\pm .006}$ 
& $5.247^{\pm .130}$ 
& $3.775^{\pm .002}$ 
& $8.102^{\pm .023}$ \\
\midrule
\multirow{4}{*}{Multi-task (Weighted)}  & Simultaneous 
& $\mathbf{0.529^{\pm .007}}$ 
& $\mathbf{0.678^{\pm .008}}$ 
& $\mathbf{0.757^{\pm .005}}$ 
& $\mathbf{4.492^{\pm .087}}$ 
& $\mathbf{3.765^{\pm .002}}$ 
& $8.065^{\pm .034}$ \\

& Reactive 
& $0.521^{\pm .009}$ 
& $0.677^{\pm .007}$ 
& $0.752^{\pm .006}$ 
& $4.736^{\pm .134}$ 
& $3.767^{\pm .002}$ 
& $8.083^{\pm .038}$ \\

& Leader--Follower 
& $0.514^{\pm .003}$ 
& $0.672^{\pm .006}$ 
& $0.751^{\pm .005}$ 
& $5.214^{\pm .138}$ 
& $3.770^{\pm .002}$ 
& $8.108^{\pm .037}$ \\

& Independent
& $0.490^{\pm .003}$ 
& $0.647^{\pm .006}$ 
& $0.731^{\pm .004}$ 
& $6.541^{\pm .147}$ 
& $3.782^{\pm .002}$ 
& $8.080^{\pm .041}$ \\
\bottomrule
\end{tabular}%
}
\end{table}

\paragraph{Inference with Different Attention Masks.}
We evaluate the flexibility of the multi-task trained model by applying different attention masks at inference. We compare two training strategies: \emph{Multi-task (Even)}, which samples the four mask types uniformly (25\% each), and \emph{Multi-task (Weighted)}, which emphasizes simultaneous interactions (60\% simultaneous, 20\% or 10\% each for others).
As shown in Table~\ref{tab:attention_mask_comparison}, the weighted multi-task model achieves the best overall performance when using the simultaneous mask (Top-1: 0.529; FID: 4.492), indicating that prioritizing the prevalent interaction pattern improves overall generation quality. The reactive mask performs comparably (Top-1: 0.521; FID: 4.736), while the leader--follower mask slightly reduces alignment but yields the highest diversity (8.108). The independent mask shows the lowest performance (Top-1: 0.490; FID: 6.541), as cross-stream coordination is disabled.

These results demonstrate that multi-task training enables controllable interaction generation via mask selection, and that emphasizing the dominant coordination pattern during training leads to stronger overall performance.

\paragraph{Computational Efficiency.} We evaluate the computational efficiency of InterCMDM on a single NVIDIA A100 GPU\@. The model contains 219.19M parameters (152.83M trainable), including 76.29M in the Temporal AutoEncoder (TAE), 76.54M in the diffusion transformer (excluding text encoder), and 66.36M in the frozen DistilBERT encoder. Peak GPU memory during inference is 566.15M.
Using 20 diffusion steps with block-wise rollout ($B{=}3$, 12 frames/block), generating a 90-frame sequence takes 1.014$\pm$0.014\,s, with a first-block latency of 0.662$\pm$0.014\,s. After the first block, streaming generation costs 50.3\,ms per 12-frame block on average, corresponding to approximately 89 FPS overall.

\section{Limitations}
\label{sec:limitations}
While InterCMDM achieves state-of-the-art results for text-conditioned two-person interaction generation, several limitations remain. First, InterCMDM is currently designed for two-person interactions; scaling to multi-person groups would require extending the dual-stream design.
Second, although cross-stream attention models coordination, the framework does not explicitly enforce physical constraints (\eg, collision avoidance or contact forces), and the mask types are manually predefined. Learning data-driven interaction masks is an important direction for future work.
Third, text conditioning relies on sentence-level embeddings from a pretrained BERT encoder, providing coarse semantic guidance. Fine-grained control over specific temporal phases or per-person action variations via an additional single-person prompt is limited. Integrating compositional text conditioning for each person could enable more precise control.

\section{Conclusion}
We present InterCMDM, a block-causal diffusion framework for autoregressive two-person interaction generation. The framework introduces three core innovations: DS-Causal-DiT, which maintains separate causal streams while modeling inter-person dependencies through unified dual-stream attention; multi-task block attention masking, which trains a unified model across diverse coordination patterns and enables controllable generation via inference-time mask selection; and block-wise diffusion, which denoises temporal blocks for long-horizon rollout directly in latent space.
Experiments on InterHuman and Inter-X demonstrate state-of-the-art performance. InterCMDM effectively combines high-fidelity diffusion synthesis with structured autoregressive temporal modeling for realistic, causally coherent human interaction generation.

%
%
\clearpage
\bibliographystyle{splncs04}
\bibliography{main}

\end{document}


\title{Supplementary Material}
\author{}
\institute{}
\maketitle

\setcounter{figure}{3}
\setcounter{table}{6}

\setcounter{page}{1}
\setcounter{section}{0  }
\renewcommand{\thesection}{\Alph{section}}

\noindent\textbf{Overview.} We provide (i) implementation details (\Sref{sec:suppl_impl}), (ii) additional quantitative results (\Sref{sec:suppl_quant}), (iii) additional qualitative results (\Sref{sec:suppl_qual}), and (iv) theoretical analyses (\Sref{sec:suppl_theory}).

\section{Implementation Details}
\label{sec:suppl_impl}

\subsection{Temporal Motion VAE}

We adopt the temporal motion VAE architecture from CMDM~\cite{yu2026causal}, which is built entirely with causal operators to preserve temporal ordering. The encoder and decoder each contain seven 1D causal convolutional layers and two causal ResNet blocks. All convolutions use kernel size 3 and stride 1 with ReLU activations. Temporal resolution is reduced by a factor of 4 using stride-2 convolutions inside the residual blocks and restored symmetrically in the decoder. The latent representation has dimensionality $d_z = 64$ per temporal step.

For training, motion sequences are segmented into 64-frame clips with a sliding window of 10 frames. The VAE is trained with AdamW using a learning rate of $2{\times}10^{-4}$, batch size 256, and a learning-rate schedule that decays by 0.1 twice over 50 epochs. Training takes approximately 1 hour on a single NVIDIA A100 GPU. Gradient clipping with a maximum norm of 1.0 is applied to stabilize training.

The model is trained using a standard variational objective applied independently to both persons:
\begin{equation}
\mathcal{L}_{\text{VAE}}
=
\sum_{i=1}^{2}
\left(
\lambda_{\text{rec}} \|\mathbf{x}^{(i)} - \hat{\mathbf{x}}^{(i)}\|^2
+
\lambda_{\text{KL}} D_{\text{KL}}
\big(
q_\phi(\mathbf{z}^{(i)}|\mathbf{x}^{(i)})
\,\|\, p(\mathbf{z}^{(i)})
\big)
\right).
\end{equation}

\subsection{DS-Causal-DiT}

The DS-Causal-DiT serves as a lightweight Transformer-based denoiser composed of 8 layers, each with 4 attention heads and a hidden dimension of 512. Text conditioning is performed via word-level embeddings extracted from the input prompt using a pretrained DistilBERT encoder~\cite{sanh2019distilbert}, which are incorporated through cross-attention.

To encode temporal structure, we incorporate Rotary Position Embeddings (RoPE)~\cite{su2024roformer} to the query and key projections in the attention, enabling relative positional modeling across time. Diffusion timesteps are injected using Adaptive Layer Normalization (AdaLN)~\cite{peebles2023scalable}, which modulates each layer with block-wise timestep information $k_b$. This design allows the model to integrate temporal noise levels directly into the denoising process while maintaining stability over long horizons.

During training, motion sequences from InterHuman and Inter-X are either clipped or padded to fixed lengths of 300 and 156 frames, respectively. The streams of Person~1 and Person~2 are randomly shuffled during training. DS-Causal-DiT is trained for 500 epochs with a batch size of 64 using the AdamW optimizer. Training takes approximately 8 hours on a single NVIDIA A100 GPU. At inference, we use 50 diffusion steps and adopt a block-wise denoising schedule from~\cite{yu2026causal}, where the next block begins denoising two steps after the previous block.

\subsection{Unified Dual-Stream Attention}
Specifically, given hidden states $\mathbf{h}_1^{(\ell-1)}, \mathbf{h}_2^{(\ell-1)} \in \mathbb{R}^{T' \times d}$, we first compute per-stream projections:
\begin{equation}
\mathbf{Q}_i = \mathbf{h}_i^{(\ell-1)} \mathbf{W}_Q^i, \quad
\mathbf{K}_i = \mathbf{h}_i^{(\ell-1)} \mathbf{W}_K^i, \quad
\mathbf{V}_i = \mathbf{h}_i^{(\ell-1)} \mathbf{W}_V^i, \quad i \in \{1,2\},
\end{equation}
where $\mathbf{W}_Q^i, \mathbf{W}_K^i, \mathbf{W}_V^i \in \mathbb{R}^{d \times d_h}$ are learned projection matrices. These are then concatenated along the sequence dimension:
\begin{equation}
\mathbf{Q} = [\mathbf{Q}_1; \mathbf{Q}_2], \quad
\mathbf{K} = [\mathbf{K}_1; \mathbf{K}_2], \quad
\mathbf{V} = [\mathbf{V}_1; \mathbf{V}_2] \in \mathbb{R}^{2T' \times d_h}.
\end{equation}
The unified dual-stream attention is then computed as standard masked multi-head attention:
\begin{equation}
\mathrm{DualStreamAttn}([\mathbf{h}_1; \mathbf{h}_2], \mathcal{M}) = \mathrm{softmax}\left(\frac{\mathbf{Q}\mathbf{K}^\top}{\sqrt{d_h}} + \mathcal{M}\right) \mathbf{V},
\end{equation}
where $\mathcal{M} \in \mathbb{R}^{2T' \times 2T'}$ is the structured attention mask with $\mathcal{M}[i,j] = -\infty$ for disallowed attention pairs and $\mathcal{M}[i,j] = 0$ otherwise. The output is then split back into $[\tilde{\mathbf{h}}_1^{(\ell)}; \tilde{\mathbf{h}}_2^{(\ell)}]$ along the sequence dimension.

\subsection{Multi-Task Block Attention Masking}
Formally, let tokens be indexed by $(p, b, t)$ where $p \in \{1,2\}$ is the person index, $b \in \{1,\dots,N_B\}$ is the block index, and $t \in \{1,\dots,B\}$ is the position within the block. The mask $\mathcal{M}$ determines whether token $(p_i, b_i, t_i)$ can attend to token $(p_j, b_j, t_j)$. For causal generation, we apply the \textbf{block-causal constraint} $b_j \le b_i$ (no future block access) to each stream’s self-attention and allow full attention within a block since we denoise blocks jointly. Cross-stream permissions are then specified by the particular mask type (e.g., full access in reactive and local-window access in leader--follower).

\begin{itemize}
\item \textbf{Simultaneous:} $\mathcal{M}[(p_i,b_i,t_i), (p_j,b_j,t_j)] = 0$ if $b_j \le b_i$; otherwise $-\infty$. Both streams have symmetric causal access.
\item \textbf{Reactive:} As reaction generation task assumes that one actor is given and fully observed, one person (reactor) is generated block-causally while conditioning on the other person’s \emph{complete} motion. Let $r\in\{1,2\}$ denote the reactor index (default $r=2$, but roles can be swapped) and $o\defeq 3-r$ the observed index. The reactor has full access to the observed stream: $\mathcal{M}[(r,b_i,t_i),(o,b_j,t_j)]=0$ for all $(b_j,t_j)$, while its self-attention remains block-causal ($b_j\le b_i$). The observed stream uses full self-attention. We consider both a cross-stream version (observed $\rightarrow$ reactor is block-causal) as the default setting and a one-way version that disables this reverse cross-stream attention.
\item \textbf{Leader--Follower:} Both self-attentions are block-causal. Person~1 (leader) does not attend to Person~2. Person~2 (follower) attends to a finite \emph{local window} of the leader around the corresponding timestep: letting $\tau(b,t)\defeq (b-1)B+t$ be the within-stream token index and $\Delta$ be the window parameter, $(2,b_i,t_i)$ may attend to $(1,b_j,t_j)$ only if $\max(1,\tau_i-\Delta+1)\le \tau_j\le \min(N_B B,\tau_i+\Delta-1)$, where $\tau_i\defeq\tau(b_i,t_i)$ and $\tau_j\defeq\tau(b_j,t_j)$.
\item \textbf{Independent:} Cross-stream attention is disabled: $\mathcal{M}[(p_i,b_i,t_i), (p_j,b_j,t_j)] = -\infty$ if $p_i \ne p_j$. Each person attends only to its own block-causal history ($b_j \le b_i$ within the same stream).
\end{itemize}
During training, the reactor identity $r$ in the reactive mask is sampled uniformly from $\{1,2\}$, and the leader--follower window parameter $\Delta$ is sampled uniformly from a discrete range $\Delta \in \{10, \dots, 30\}$.

\section{Additional Quantitative Results}
\label{sec:suppl_quant}
\subsection{Causality, Downsampling, and Latent Dimensionality of Temporal Motion VAE}
We ablate the Temporal Motion VAE along three factors: causality, temporal downsampling rate, and latent dimensionality. We evaluate both reconstruction quality and downstream text-to-motion generation in the latent space. As shown in Table~\ref{tab:tae_ablation}, more aggressive temporal compression substantially degrades reconstruction quality. For example, increasing the downsampling rate from 4$\times$ to 8$\times$ raises MPJPE from 0.025 to 0.173 when $d_z{=}64$, while producing only marginal changes in DiT generation metrics. Increasing the latent dimensionality improves reconstruction, but does not necessarily improve generation quality; for instance, $d_z{=}128$ yields a higher generation FID than $d_z\in\{32,64\}$. We also find that VAE causality is important for causal generation. The non-causal VAE reduces R-Top1 from 0.529 to 0.475 and worsens generation FID from 4.492 to 4.739. Based on these trade-offs, we use a causal VAE with 4$\times$ downsampling and $d_z{=}64$ in all experiments.

\begin{table}[t]
\caption{Temporal VAE ablations on InterHuman. We compare downsampling rates, latent dimensions, and VAE causality. Reconstruction metrics evaluate VAE reconstruction quality, while generation metrics evaluate interaction generation. The average is reported over 10 runs with 95\% confidence intervals.}
\label{tab:tae_ablation}
\centering
\resizebox{\linewidth}{!}{
\begin{tabular}{c|cc|ccc|cc|c}
\toprule
\multirow{2}{*}{\shortstack{VAE setting\\(downsampling / latent dim.)}}
& \multicolumn{2}{c|}{Reconstruction Metrics}
& \multicolumn{6}{c}{Generation Metrics} \\
\cline{2-3} \cline{4-9}
& MPJPE$\downarrow$ & FID$\downarrow$ & R-Top1$\uparrow$ & R-Top2$\uparrow$ & R-Top3$\uparrow$ & FID$\downarrow$ & MM Dist$\downarrow$ & Diversity$\rightarrow$ \\
\midrule
$4\times$ / 32  & $0.036^{\pm .030}$ & $0.644^{\pm .021}$ & $0.525^{\pm .006}$ & $0.675^{\pm .006}$ & $0.755^{\pm .006}$ & $\mathbf{4.322^{\pm .069}}$ & $3.769^{\pm .001}$ & $8.040^{\pm .034}$ \\
\rowcolor[gray]{0.90} $4\times$ / 64  & $0.025^{\pm .020}$ & $\mathbf{0.594^{\pm .018}}$ & $\mathbf{0.529^{\pm .007}}$ & $0.678^{\pm .008}$ & $0.757^{\pm .005}$ & $4.492^{\pm .087}$ & $\mathbf{3.765^{\pm .002}}$ & $8.065^{\pm .034}$ \\
$4\times$ / 128 & $\mathbf{0.020^{\pm .018}}$ & $0.604^{\pm .017}$ & $0.506^{\pm .006}$ & $0.663^{\pm .005}$ & $0.743^{\pm .004}$ & $5.659^{\pm .143}$ & $3.775^{\pm .001}$ & $8.125^{\pm .045}$ \\
\midrule
$8\times$ / 32  & $0.178^{\pm .161}$ & $1.531^{\pm .021}$ & $0.520^{\pm .008}$ & $0.678^{\pm .010}$ & $0.756^{\pm .008}$ & $4.333^{\pm .083}$ & $3.768^{\pm .002}$ & $\mathbf{7.930^{\pm .036}}$ \\
$8\times$ / 64  & $0.173^{\pm .162}$ & $1.464^{\pm .020}$ & $0.528^{\pm .004}$ & $\mathbf{0.685^{\pm .006}}$ & $\mathbf{0.760^{\pm .005}}$ & $4.366^{\pm .076}$ & $3.768^{\pm .002}$ & $8.002^{\pm .036}$ \\
$8\times$ / 128 & $0.074^{\pm .065}$ & $1.423^{\pm .015}$ & $0.514^{\pm .008}$ & $0.673^{\pm .004}$ & $0.750^{\pm .004}$ & $5.048^{\pm .097}$ & $3.773^{\pm .001}$ & $8.051^{\pm .036}$ \\
\midrule
Non-causal VAE & $0.028^{\pm .023}$ & $0.601^{\pm .019}$ & $0.475^{\pm .007}$ & $0.635^{\pm .005}$ & $0.709^{\pm .005}$ & $4.739^{\pm .125}$ & $3.794^{\pm .001}$ & $7.962^{\pm .030}$ \\
\bottomrule
\end{tabular}
}
\end{table}

\subsection{Model Size of DS-Causal-DiT}
We ablate the denoiser capacity by scaling DS-Causal-DiT in terms of the number of attention heads (H), layers (L), and hidden dimensions (D), ranging from a small model (4 layers, 39M parameters) to an extra-large model (16 layers, 607M parameters), while keeping the dual-stream block-wise causal attention and training setup fixed. As shown in Table~\ref{tab:model_size_ablation}, increasing the size from Small to Base substantially improves both text--motion alignment and generation quality (\eg, R-Precision Top-1: 0.487$\rightarrow$0.529; FID: 5.501$\rightarrow$4.492). Further scaling yields diminishing returns: Large/XLarge provide only marginal R-Precision gains (Top-1 up to 0.538) and do not consistently improve FID/MM Dist, while diversity remains comparable. Based on this accuracy--efficiency trade-off, we adopt the Base configuration (8 layers, hidden size 512, 77M parameters; excluding the text encoder) in all experiments.

\begin{table}[t]
\caption{Performance comparison across different model scales on InterHuman. The average is reported over 10 runs with 95\% confidence intervals. Model parameters exclude BERT encoder.}
\label{tab:model_size_ablation}
\centering
\resizebox{\linewidth}{!}{
\begin{tabular}{l|c|c|ccc|cc|c}
\toprule
\multirow{2}{*}{Model}
& \multirow{2}{*}{Configuration}
& \multirow{2}{*}{\#Params}
& \multicolumn{3}{c|}{R-Precision$\uparrow$}
& \multirow{2}{*}{FID$\downarrow$}
& \multirow{2}{*}{MM Dist$\downarrow$}
& \multirow{2}{*}{Diversity$\rightarrow$} \\
\cline{4-6}
& & & Top-1 & Top-2 & Top-3 & & & \\
\midrule
Small & H2, L4, D512 & 39M & $0.487^{\pm .009}$ & $0.646^{\pm .007}$ & $0.729^{\pm .007}$ & $5.501^{\pm .106}$ & $3.780^{\pm .002}$ & $8.106^{\pm .046}$ \\
\rowcolor[gray]{0.90} Base & H4, L8, D512 & 77M & $0.529^{\pm .007}$ & $0.678^{\pm .008}$ & $0.757^{\pm .005}$ & $\mathbf{4.492^{\pm .087}}$ & $3.765^{\pm .002}$ & $8.065^{\pm .034}$ \\
Large & H6, L12, D768 & 257M & $0.533^{\pm .011}$ & $0.680^{\pm .007}$ & $0.753^{\pm .008}$ & $4.764^{\pm .104}$ & $3.763^{\pm .002}$ & $\mathbf{7.975^{\pm .041}}$ \\
XLarge & H8, L16, D1024 & 607M & $\mathbf{0.538^{\pm .008}}$ & $\mathbf{0.684^{\pm .009}}$ & $\mathbf{0.757^{\pm .007}}$ & $4.706^{\pm .072}$ & $\mathbf{3.761^{\pm .002}}$ & $8.001^{\pm .055}$ \\
\bottomrule
\end{tabular}
}
\end{table}

\subsection{Variations of Reactive Mask}
In the reactive setting, the reactor is generated block-causally while conditioning on the observed person’s \emph{complete} motion. Under our unified dual-stream attention, there is an additional design choice: whether to allow the observed stream to attend to the reactor. Our default \textbf{Reactive} mask keeps this reverse cross-stream attention block-causal, whereas \textbf{Reactive*} disables it (one-way), treating the observed stream as a fixed context.

Table~\ref{tab:reactive_ablation} reports results on InterHuman. For standard interaction generation (inference: simultaneous), adding Reactive* as an extra training mask has negligible impact (Top-1: 0.529$\rightarrow$0.527; FID: 4.492$\rightarrow$4.544). For reaction generation (Person~1 given, Person~2 generated), training only with Reactive or Reactive* yields noticeably worse text alignment than multi-task training (Top-1: 0.451/0.450 vs 0.484), while Reactive* slightly improves FID over Reactive (2.216$\rightarrow$2.160). Overall, the multi-task model generalizes best to reactive inference, achieving the lowest FID (2.121) with strong alignment. Adding Reactive* to multi-task marginally improves R-Precision (Top-1: 0.484$\rightarrow$0.487) but degrades FID (2.121$\rightarrow$2.387), so we use the standard Reactive mask in multi-task training in all experiments.

\begin{table}[t]
\centering
\caption{Ablation of reactive-mask variants on InterHuman. Top: standard interaction generation evaluated with the simultaneous mask. Bottom: reaction generation, where Person~1 is provided as ground truth and Person~2 is generated as the response.}
\label{tab:reactive_ablation}
\resizebox{\columnwidth}{!}{%
\begin{tabular}{lc|ccc|cc|c}
\toprule
\multirow{2}{*}{Training Mask} 
& \multirow{2}{*}{Inference Mask} 
& \multicolumn{3}{c|}{R-Precision$\uparrow$} 
& \multirow{2}{*}{FID$\downarrow$} 
& \multirow{2}{*}{MM Dist$\downarrow$} 
& \multirow{2}{*}{Diversity$\rightarrow$} \\
\cline{3-5}
& & Top-1 & Top-2 & Top-3 & & & \\
\midrule
\multicolumn{8}{c}{\textit{Interaction Generation}} \\ 
Multi-task & \multirow{2}{*}{Simultaneous} 
& $\mathbf{0.529^{\pm .007}}$ 
& $\mathbf{0.678^{\pm .008}}$ 
& $\mathbf{0.757^{\pm .005}}$ 
& $\mathbf{4.492^{\pm .087}}$ 
& $\mathbf{3.765^{\pm .002}}$ 
& $\mathbf{8.065^{\pm .034}}$ \\

Multi-task (w/ Reactive*) & 
& $0.527^{\pm .001}$ 
& $0.678^{\pm .006}$ 
& $0.751^{\pm .007}$ 
& $4.544^{\pm .098}$ 
& $3.767^{\pm .001}$ 
& $8.065^{\pm .040}$ \\
\midrule
\multicolumn{8}{c}{\textit{Reaction Generation: Person 1 (GT) and Person 2 (Reactor)}} \\ 
Reactive & \multirow{4}{*}{Reactive} 
& $0.451^{\pm .006}$ & $0.612^{\pm .006}$ & $0.698^{\pm .006}$
& $2.216^{\pm .049}$ & $3.778^{\pm .002}$ & $7.843^{\pm .037}$ \\
Reactive* & 
& $0.450^{\pm .008}$ & $0.613^{\pm .007}$ & $0.701^{\pm .007}$
& $2.160^{\pm .058}$ & $3.779^{\pm .002}$ & $7.806^{\pm .022}$  \\
Multi-task & 
& $0.484^{\pm .005}$ & $0.642^{\pm .006}$ & $0.727^{\pm .006}$
& $\mathbf{2.121^{\pm .033}}$ & $3.771^{\pm .002}$ & $7.751^{\pm .019}$ \\
Multi-task (w/ Reactive*) & 
& $\mathbf{0.487^{\pm .009}}$ & $\mathbf{0.649^{\pm .007}}$ & $\mathbf{0.733^{\pm .006}}$
& $2.387^{\pm .051}$ & $\mathbf{3.771^{\pm .001}}$ & $\mathbf{7.743^{\pm .026}}$ \\ 
\bottomrule
\end{tabular}
}
\end{table}

\section{Additional Qualitative Results}
\label{sec:suppl_qual}

\begin{figure*}[!t]
    \centering
    \includegraphics[width=0.75\linewidth]{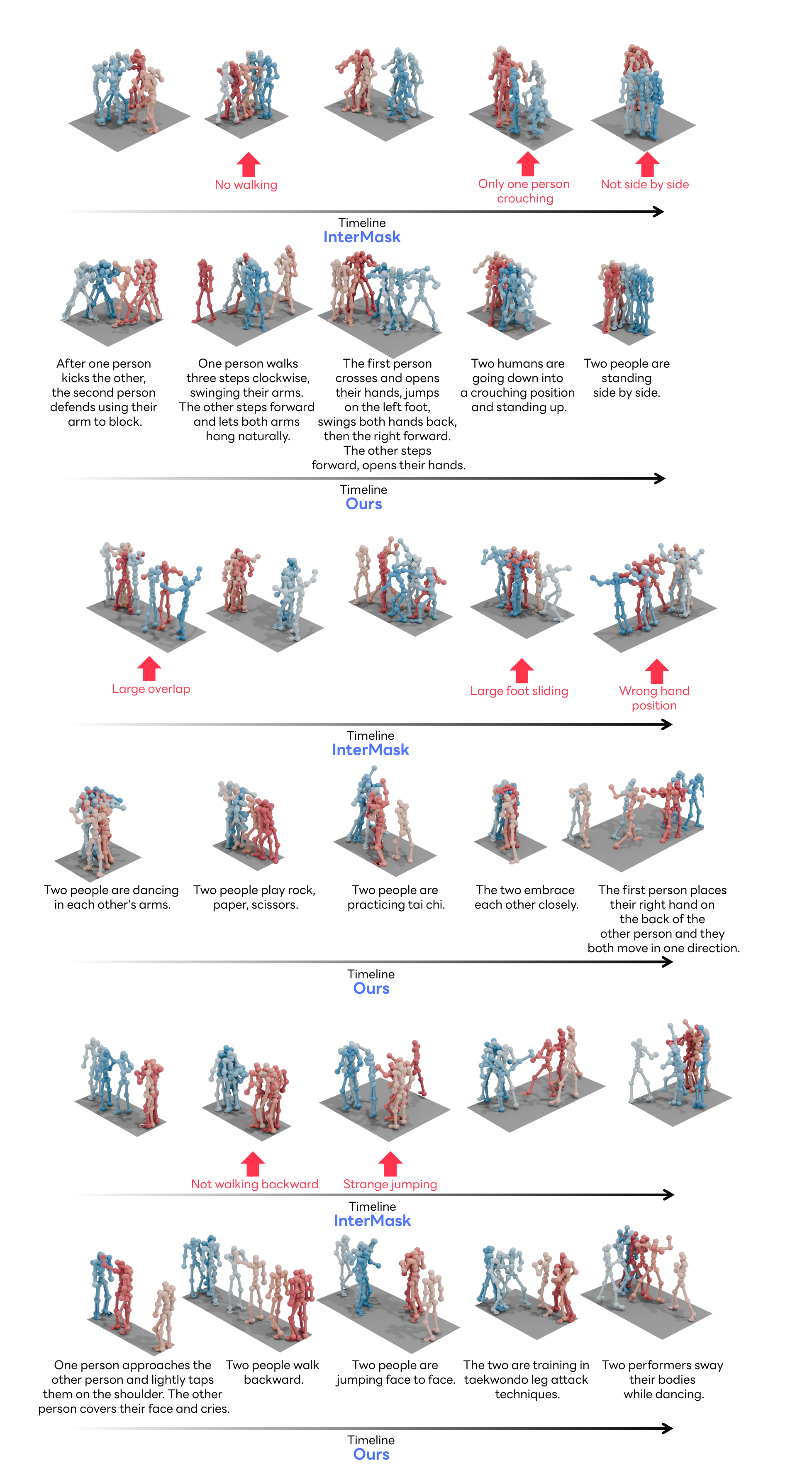}
    \vspace{-5pt}
    \caption{Additional qualitative comparison of long-horizon motion generation between InterCMDM and InterMask. Our results demonstrate more continuous and coherent interactions. For clarity, each long sequence is divided into shorter segments aligned with their corresponding captions. Please refer to the supplementary videos for full-length sequences.}
    \vspace{-10pt}
    \label{fig:qualitative_2}
\end{figure*}

\begin{figure*}[!t]
    \centering
    \includegraphics[width=0.95\linewidth]{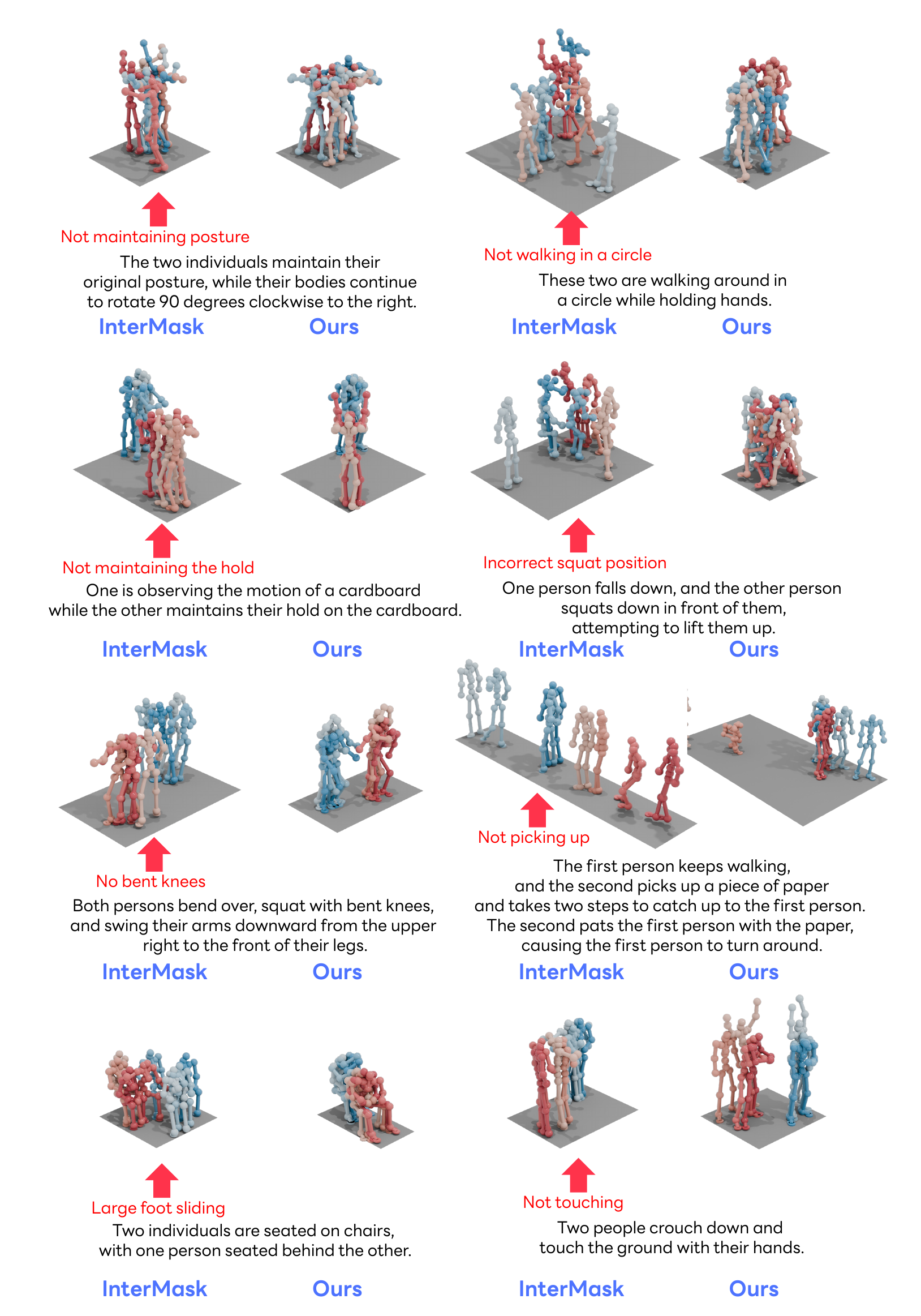}
    \vspace{-5pt}
    \caption{Additional qualitative comparison of interaction generation between InterCMDM and InterMask. Our method demonstrates better capture of fine-grained textual semantics and maintains more natural body articulation compared to previous methods. Please refer to the supplementary videos for clearer visualization.}
    \vspace{-10pt}
    \label{fig:qualitative_3}
\end{figure*}

To further demonstrate the effectiveness of InterCMDM, we provide additional qualitative comparisons against InterMask on both \emph{long-horizon} and \emph{standard} interaction generation.
\Fref{fig:qualitative_2} shows additional long-horizon examples. InterCMDM maintains coherent inter-person coordination across many caption segments, preserving mutual facing, relative distance, and contact consistency (\eg, stable hand/arm coupling in greeting-like actions) while producing smooth transitions between segments.
In contrast, InterMask exhibits accumulated drift over time, which can manifest as foot sliding, spatial overlap, and inconsistent action--reaction patterns across successive segments.

\Fref{fig:qualitative_3} presents additional short-horizon interaction generation results.
InterCMDM better captures fine-grained textual semantics and produces more natural body articulation, especially in cases requiring precise role-aware coordinations (who initiates, who responds).
These examples highlight how block-causal diffusion with unified dual-stream attention improves both temporal coherence and semantic consistency for human interaction generation.

\textbf{\textit{Please refer to the supplementary videos on the demo page for full-length visualizations.}}

\section{Theoretical Analyses}
\label{sec:suppl_theory}

This section provides theoretical analyses of our (i) \textbf{dual-stream network} design and (ii) \textbf{multi-task attention masks}. We formalize how the architecture and mask determine \emph{information flow} (which tokens can influence which predictions), and why (a) \emph{dual-stream parameterization} is strictly more expressive than a \emph{single-stream concatenation} baseline, and (b) \emph{multi-task masking} is preferable to training with a single fixed attention pattern when we want both controllability and robustness.

\subsection{Dual-Stream Network vs.\ Single-Stream Concatenation}
\label{sec:suppl_dual_vs_single}

\paragraph{Single-stream baseline (concatenation with shared projections).}
A common design for two-person modeling concatenates both people into one sequence and applies standard masked self-attention with \emph{shared} projections across all tokens:
\begin{equation}
\mathrm{Attn}_{\text{single}}(\mathbf{H};\mathcal{M})
=
\mathrm{softmax}\!\left(\frac{\mathbf{H}\mathbf{W}_Q(\mathbf{H}\mathbf{W}_K)^\top}{\sqrt{d_h}}+\mathcal{M}\right)\mathbf{H}\mathbf{W}_V,
\quad
\mathbf{H}\in\mathbb{R}^{2T'\times d}.
\end{equation}
Here, $\mathbf{H}=[\mathbf{h}_1;\mathbf{h}_2]$ and $\mathbf{W}_Q,\mathbf{W}_K,\mathbf{W}_V\in\mathbb{R}^{d\times d_h}$ are shared.

\paragraph{Dual-stream as a strict generalization.}
Our unified dual-stream attention uses \emph{stream-specific} projections. Define block-diagonal matrices
\begin{equation}
\mathbf{W}_Q \defeq \mathrm{diag}(\mathbf{W}_Q^1,\mathbf{W}_Q^2),\quad
\mathbf{W}_K \defeq \mathrm{diag}(\mathbf{W}_K^1,\mathbf{W}_K^2),\quad
\mathbf{W}_V \defeq \mathrm{diag}(\mathbf{W}_V^1,\mathbf{W}_V^2),
\end{equation}
so that $\mathbf{Q}=\mathbf{H}\mathbf{W}_Q$, $\mathbf{K}=\mathbf{H}\mathbf{W}_K$, and $\mathbf{V}=\mathbf{H}\mathbf{W}_V$ reproduce the per-stream projections and concatenation. Therefore, our attention is exactly masked self-attention on $\mathbf{H}$ with mask $\mathcal{M}$, but with a \emph{block-structured parameterization}. Importantly, the single-stream baseline is recovered by the special case $\mathbf{W}_Q^1=\mathbf{W}_Q^2$, $\mathbf{W}_K^1=\mathbf{W}_K^2$, and $\mathbf{W}_V^1=\mathbf{W}_V^2$ (and identical FFN/TextAttn weights). Hence, dual-stream \emph{contains} single-stream as a special case.

\paragraph{Why dual-stream is strictly more expressive.}
Stream-specific projections decouple the similarity metrics used for self- and cross-person matching. Let
\begin{equation}
\mathbf{S} \defeq \frac{\mathbf{Q}\mathbf{K}^\top}{\sqrt{d_h}}
=
\begin{bmatrix}
\mathbf{S}_{11} & \mathbf{S}_{12}\\
\mathbf{S}_{21} & \mathbf{S}_{22}
\end{bmatrix},
\qquad
\mathbf{S}_{ij}=\frac{(\mathbf{h}_i\mathbf{W}_Q^i)(\mathbf{h}_j\mathbf{W}_K^j)^\top}{\sqrt{d_h}}.
\end{equation}
Dual-stream allows $\mathbf{S}_{11},\mathbf{S}_{22}$ (within-person) and $\mathbf{S}_{12},\mathbf{S}_{21}$ (cross-person) to be governed by different bilinear forms, capturing asymmetric interaction patterns and person-specific dynamics. In contrast, a shared-projection single-stream design ties all four blocks to the same $(\mathbf{W}_Q,\mathbf{W}_K)$, reducing flexibility.

\paragraph{Self vs.\ cross coupling is explicit.}
Let $\mathbf{A}_{\mathcal{M}}\defeq\mathrm{softmax}(\mathbf{S}+\mathcal{M})$ and partition it into blocks $\mathbf{A}_{ij}$ analogously. Then the attention output decomposes as
\begin{equation}
\begin{bmatrix}
\tilde{\mathbf{h}}_1\\
\tilde{\mathbf{h}}_2
\end{bmatrix}
=
\mathbf{A}_{\mathcal{M}}
\begin{bmatrix}
\mathbf{V}_1\\
\mathbf{V}_2
\end{bmatrix},
\qquad
\tilde{\mathbf{h}}_1=\mathbf{A}_{11}\mathbf{V}_1+\mathbf{A}_{12}\mathbf{V}_2,\quad
\tilde{\mathbf{h}}_2=\mathbf{A}_{21}\mathbf{V}_1+\mathbf{A}_{22}\mathbf{V}_2.
\end{equation}
The off-diagonal blocks $(\mathbf{A}_{12},\mathbf{A}_{21})$ are the only cross-person information pathways, and the mask $\mathcal{M}$ directly enables/disables them (e.g., $\mathbf{A}_{12}=\mathbf{A}_{21}=\mathbf{0}$ for the independent mask).

\paragraph{Proposition (symmetry breaking).}
Consider the minimal case $T'=1$ with a mask that allows mutual attention. If the two inputs are identical ($\mathbf{h}_1=\mathbf{h}_2$) but the model uses different value projections ($\mathbf{W}_V^1\ne \mathbf{W}_V^2$), then the dual-stream outputs for the two tokens can be different. A single-stream attention layer with shared $\mathbf{W}_V$ cannot produce different outputs under $\mathbf{h}_1=\mathbf{h}_2$ unless additional person-specific parameters are introduced. Thus, stream-specific projections provide a structural mechanism to model role/asymmetry without relying purely on learned person-ID conditioning.

\subsection{Mask-Induced Causal Information Flow}
\label{sec:suppl_mask_flow}

\paragraph{Mask-induced dependency set.}
For any token index $i$, define its allowed attention set
\begin{equation}
\mathcal{A}_{\mathcal{M}}(i) \defeq \{\,j \mid \mathcal{M}[i,j]=0\,\}.
\end{equation}
In a masked attention layer, the output at token $i$ is a convex combination of $\{\mathbf{V}[j]\}_{j\in \mathcal{A}_{\mathcal{M}}(i)}$ (after softmax), hence it depends only on the values of tokens in $\mathcal{A}_{\mathcal{M}}(i)$. The mask therefore defines a directed graph over tokens with edges $j\rightarrow i$ whenever $j\in\mathcal{A}_{\mathcal{M}}(i)$.

\paragraph{Lemma (Block-causal information flow).}
Assume $\mathcal{M}$ enforces the block-causal constraint $b_j \le b_i$ for the tokens being predicted. Then for any layer $\ell$, the representation of any token $(p,b,t)$ after $\ell$ Dual-Stream blocks depends only on tokens from blocks $\le b$ (from either person, depending on cross-stream permissions in $\mathcal{M}$). Reactive masks additionally allow the reactor to attend to the observed person’s full sequence by design.

\emph{Proof (sketch).} At $\ell=1$, masked attention restricts aggregation to $b_j \le b$. Residual connections, per-token FFNs, and text cross-attention do not introduce access to future blocks. Inductively, if layer $\ell-1$ representations depend only on blocks $\le b$, then layer $\ell$ cannot introduce dependence on $b'>b$ because attention is still masked by $b_j \le b$. \qed

\paragraph{Remark (propagation depth).}
Stacking $L$ masked blocks allows information to propagate along paths of length up to $L$ in the mask-induced token graph, while still respecting block causality. This clarifies how interaction cues (e.g., a leader’s early motion) can influence a follower several blocks later, depending on both $L$ and the mask.

\subsection{Multi-Task Masks vs.\ Single-Task Attention}
\label{sec:suppl_mask_theory}

Let $\boldsymbol{\mu}_\theta(\cdot;\mathcal{M})$ denote the DS-Causal-DiT denoiser under mask $\mathcal{M}$ (predicting noise residuals per token/block). Because the only coupling between the two people inside a Dual-Stream block occurs through the \emph{off-diagonal} attention blocks ($\mathbf{S}_{12},\mathbf{S}_{21}$), each mask $\mathcal{M}$ induces a distinct set of cross-person conditional dependencies and, equivalently, a distinct conditional factorization of the block-wise denoising problem.

\paragraph{Mask-induced per-block conditioning sets.}
Let $\mathcal{C}^{(p)}_{\mathcal{M}}(b)$ denote the set of tokens from which person $p$ at block $b$ can receive information under $\mathcal{M}$ (including within-person causal history and any allowed cross-person history). Then the denoiser for person $p$ at block $b$ is restricted to the form
\begin{equation}
\boldsymbol{\mu}_\theta^{(p,b)}=\boldsymbol{\mu}_\theta^{(p,b)}\!\left(\mathbf{z}_{\mathbf{k}};\mathcal{C}^{(p)}_{\mathcal{M}}(b)\right),
\end{equation}
where $\mathbf{z}_{\mathbf{k}}$ denotes the collection of noisy latent blocks. The four masks correspond to the following canonical conditioning patterns (omitting within-block indices for brevity):
\begin{itemize}
\item \textbf{Simultaneous (symmetric)}:
\begin{equation*}
\begin{aligned}
\mathcal{C}^{(1)}_{\mathcal{M}}(b) &= \mathcal{C}^{(2)}_{\mathcal{M}}(b) \\
&= \{\mathbf{z}^{(1,b')},\mathbf{z}^{(2,b')}\}_{b'\le b}.
\end{aligned}
\end{equation*}
\item \textbf{Reactive (asymmetric):} Let $r$ be the reactor and $o$ the observed stream. The reactor conditions on its own history up to $b$ and the observed person’s \emph{entire} sequence:
\begin{equation*}
\mathcal{C}^{(r)}_{\mathcal{M}}(b)=\{\mathbf{z}^{(r,b')}\}_{b'\le b}\cup\{\mathbf{z}^{(o,b')}\}_{b'=1}^{N_B}.
\end{equation*}
The observed stream uses full self-context over all blocks; in the cross-stream version it may additionally incorporate reactor blocks up to $b$, whereas the one-way version removes this reverse cross-stream dependency.
\item \textbf{Leader--Follower (windowed):} $\mathcal{C}^{(1)}_{\mathcal{M}}(b)=\{\mathbf{z}^{(1,b')}\}_{b'\le b}$ and $\mathcal{C}^{(2)}_{\mathcal{M}}(b)$ includes $\{\mathbf{z}^{(2,b')}\}_{b'\le b}$ plus a finite local window of leader tokens around the follower’s time indices (controlled by $\Delta$ as defined above).
\item \textbf{Independent (decoupled):} $\mathcal{C}^{(p)}_{\mathcal{M}}(b)=\{\mathbf{z}^{(p,b')}\}_{b'\le b}$ (no cross-person tokens).
\end{itemize}
Thus, changing $\mathcal{M}$ is equivalent to choosing a different, explicit cross-person dependency structure while retaining block-causal self-attention for the stream(s) generated autoregressively.

\paragraph{Optimal denoiser under a mask.}
Under the squared error diffusion objective, the Bayes-optimal predictor for each $(p,b)$ is the conditional expectation given the information permitted by the mask:
\begin{equation}
\boldsymbol{\mu}_*^{(p,b)}(\cdot;\mathcal{M})
=
\mathbb{E}\!\left[\boldsymbol{\epsilon}^{(p,b)} \,\middle|\, \mathcal{C}^{(p)}_{\mathcal{M}}(b)\right].
\end{equation}
Different masks therefore correspond to different target conditional expectations (different conditioning sets). Training with a single fixed mask learns only one such conditional, whereas multi-task masking explicitly trains the same parameters to approximate this family $\{\boldsymbol{\mu}_*^{(p,b)}(\cdot;\mathcal{M}_m)\}_m$.

\paragraph{Multi-task masking objective (amortized over masks).}
During training, we sample a mask type $m$ (with corresponding mask $\mathcal{M}_m$) from a distribution $\pi(m)$. The training objective becomes an expectation over both diffusion noise and masks:
\begin{equation}
\mathcal{L}(\theta)
=
\mathbb{E}_{m \sim \pi}
\left[
\mathcal{L}_m(\theta)
\right],
\qquad
\mathcal{L}_m(\theta)
\defeq
\mathbb{E}_{\mathbf{k}, \boldsymbol{\epsilon}}
\left[
\sum_{i=1}^{2}\sum_{b=1}^{N_B}
\left\|
\boldsymbol{\epsilon}^{(i,b)} - \boldsymbol{\mu}_\theta\!\left(\cdot;\mathcal{M}_m\right)
\right\|_2^2
\right].
\end{equation}
Single-task attention is the degenerate case where $\pi(m)$ places all its mass on one mask. In that setting, switching to a different mask at inference changes the model’s effective computation graph (removing or adding attention edges) and leads to a form of \emph{structural distribution shift}. Multi-task masking eliminates this mismatch by explicitly training the same parameters $\theta$ under multiple attention graphs.

\paragraph{Advantages over single-task training.}
Multi-task masking provides: (i) \textbf{controllability}---a single $\theta$ supports multiple coordination regimes by selecting $\mathcal{M}$ at inference; (ii) \textbf{structural regularization}---randomly sampling masks acts like structured edge dropout, discouraging over-reliance on any single cross-person pathway and improving robustness; and (iii) \textbf{amortized modeling of multiple factorizations}---instead of training separate parameters $\{\theta_m\}$ for each dependency structure, we share representations across masks while keeping the dependency constraints explicit and verifiable through $\mathcal{M}$.

\bibliographystyle{splncs04}
\bibliography{main}